Research Paper

# Attention-driven next-best-view planning for efficient reconstruction of plants and targeted plant parts

Akshay K. Burusa *, Eldert J. van Henten, Gert Kootstra

*Agricultural Biosystems Engineering, Wageningen University & Research, Wageningen, the Netherlands*

A B S T R A C T

Robots in tomato greenhouses need to perceive the plant and plant parts accurately to automate monitoring, harvesting, and de-leafing tasks. Existing perception systems struggle with the high levels of occlusion in plants and often result in poor perception accuracy. One reason for this is because they use fixed cameras or predefined camera movements. Next-best-view (NBV) planning presents an alternate approach, in which the camera viewpoints are reasoned and strategically planned such that the perception accuracy is improved. However, existing NBV-planning algorithms are agnostic to the task-at-hand and give equal importance to all the plant parts. This strategy is inefficient for greenhouse tasks that require targeted perception of specific plant parts, such as the perception of leaf nodes for de-leafing. To improve targeted perception in complex greenhouse environments, NBV planning algorithms need an attention mechanism to focus on the task-relevant plant parts. In this paper, the role of attention in improving targeted perception using an attention-driven NBV planning strategy was investigated. Through simulation experiments using plants with high levels of occlusion and structural complexity, it was shown that focusing attention on task-relevant plant parts can significantly improve the speed and accuracy of 3D reconstruction. Further, with real-world experiments, it was shown that these benefits extend to complex greenhouse conditions with natural variation and occlusion, natural illumination, sensor noise, and uncertainty in camera poses. The results clearly indicate that using attention-driven NBV planning in greenhouses can significantly improve the efficiency of perception and enhance the performance of robotic systems in greenhouse crop production.

Nomenclature

| | |
|---|---|
| 3D | Three dimensional |
| 6D | Six dimensional |
| CPU | Central processing unit |
| DoF | Degrees of freedom |
| GHz | Gigahertz |
| LED | Light emitting diode |
| LiDAR | Light detection and ranging |
| MB | Megabyte |
| NBV | Next-best view |
| RGB-D | Red Green Blue colour and depth |
| ROI | Region of interest |
| ROS | Robotic operating system |

## 1. Introduction

Tomatoes are one of the most consumed vegetables in the world today (Beed et al., 2021). As the world population increases, the demand for tomatoes is estimated to grow even further. To meet this rising demand, tomato growers aim to scale up production, but they are limited by the high costs of labour and an ageing and diminishing labour force (White, 2012). The dual problem of increasing demand and limited availability of labour has triggered an interest in robotic solutions in tomato greenhouses. With the help of robots, the labour-intensive tasks in a greenhouse can be automated. This can reduce the dependence on human labour and significantly scale up tomato production.

To operate in a tomato greenhouse, robots should be able to perceive the plants accurately. This includes observing the geometric structure of the plant, identifying different plant parts, and estimating their properties, such as position and orientation. Perceiving such information allows robots to successfully perform downstream tasks such as monitoring the plant growth, removing leaves, and harvesting ripe tomatoes without causing any damage to the plant. With recent advancements in computer vision, plants can be visually perceived by the process of 3D reconstruction, in which an RGB-D camera is used to capture the shape and appearance of the plants and make sense of their geometric structure. However, the lack of an accurate and robust visual perception system remains to be one of the biggest challenges for greenhouse robots and significantly limits their commercial success (Shamshiri et al., 2018).

* Corresponding author.
*E-mail address:* akshaykumar.burusa@wur.nl (A.K. Burusa).

Visual perception in a greenhouse environment is extremely challenging due to several factors. First, plants exhibit natural variation in their physical and environmental properties, such as size, shape, colour, illumination, and background. Most perception systems are developed to work under a certain range of conditions and hence generalise poorly to variation. Second, plant parts often occlude each other, which leads to incomplete information about the plant, as shown in Fig. 1. For example, tomatoes can be hidden from the robot's view by leaves or other tomatoes. Most perception systems do not have explicit strategies to deal with occlusion and hence operate inefficiently under partial information (Bac et al., 2014, 2017). Third, each greenhouse task has different perception requirements. For example, monitoring the plant growth requires perception of the complete plant, while de-leafing and harvesting tasks require the perception of only the leaf and fruit nodes attached to the main stem. Most perception systems are not flexible enough to handle a wide range of perception requirements and hence do not adapt well to the task-at-hand. Due to these challenges, the large-scale deployment of robots in tomato greenhouses still has a long way to go (Kootstra et al., 2021). To ensure robust operation of greenhouse robots, it is essential that these challenges are addressed.

Variation and occlusion in plants can be handled by using multiple viewpoints to gather more information, as demonstrated by Hemming et al. (2014), who showed a 35% increase in fruit detectability when multiple viewpoints were used instead of a single viewpoint. By moving to a new viewpoint, robots gain new information that is otherwise unavailable. However, the selection of viewpoints is crucial to efficiently gather the relevant information, that is, using the least number of viewpoints. Most multi-view 3D reconstruction approaches in practice follow the paradigm of passive perception, wherein the selection of viewpoints is predefined, usually chosen by human operators based on their intuition or reasoning about which viewpoints will capture the most information about the plant (Boogaard et al., 2020). While such passive approaches work well to some extent, they often suffer from sub-optimal viewpoint selection, requiring more viewpoints to gather the necessary information. Furthermore, they require manual readjustment of viewpoints for novel plants or scenes.

Recently, there has been a growing interest in the paradigm of active vision (Bircher et al., 2016; Devrim Kaba et al., 2017; Isler et al., 2016; Kriegel et al., 2015), first proposed by Bajcsy (1988) and Aloimonos et al. (1988). Under this paradigm, a robot automatically selects camera viewpoints to capture novel information. The view selection is based on the perception objective and previously acquired information about an object or scene. In contrast to passive approaches, active vision reasons about the observed and unobserved parts of space, which allows it to generalise well to novel objects and scenes. Recent literature demonstrates that active vision can significantly improve the quality of plant perception and reduce the number of required viewpoints compared to passive perception (Chen et al., 2024; Gibbs et al., 2018, 2019).

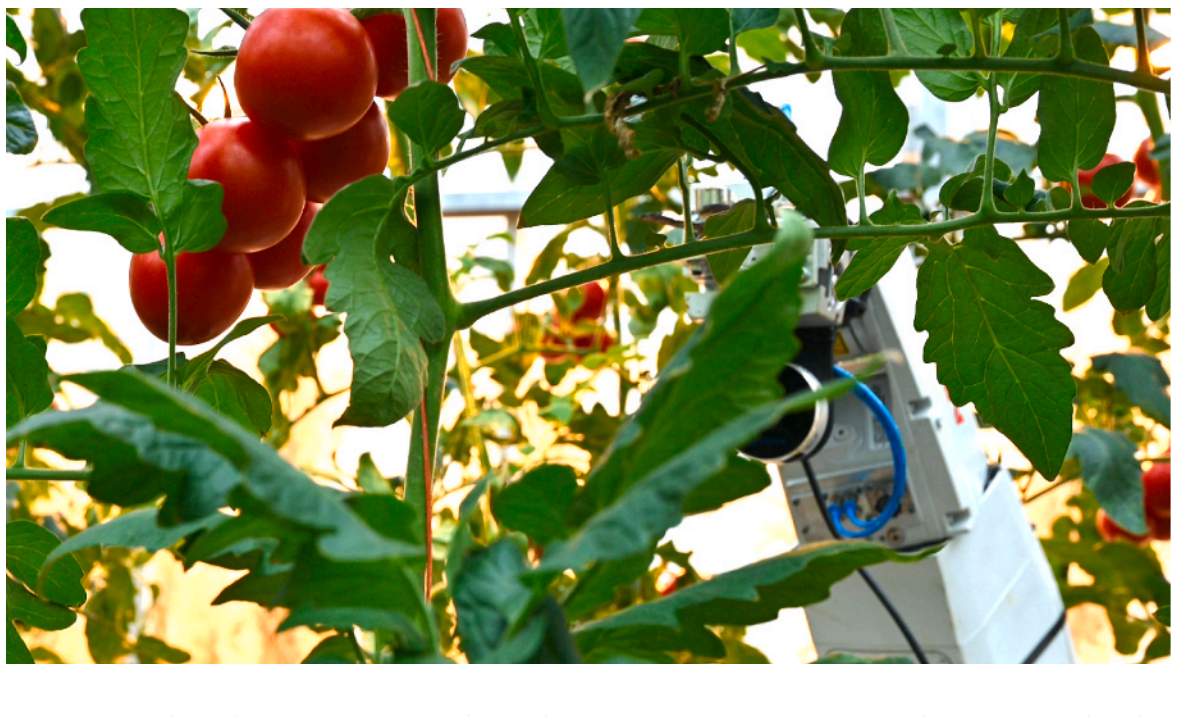

**Fig. 1.** Example of variation and occlusion in tomato greenhouses, which make the visual perception of plant parts challenging.

Current active-vision algorithms are primarily designed for complete object or scene reconstruction (Bircher et al., 2016; Devrim Kaba et al., 2017; Isler et al., 2016; Kriegel et al., 2015). They use all the information acquired from the previous viewpoints to select the next view. In the context of plant reconstruction, such a strategy gives equal importance to all plant parts and is only useful for tasks in which perception of the whole plant is relevant, such as growth monitoring. However, for tasks like de-leafing and harvesting, which require the perception of specific plant parts, current active-vision strategies might be extremely inefficient. Hence, we hypothesise that for greenhouse tasks, a strategy that can focus on the most relevant plant parts according to the task-at-hand, i.e. an attention mechanism, is essential. A recent study by Zaenker et al. (2021) proposed the use of labelled regions of interest along with active vision to improve the perception of fruits in sweet pepper plants and showed significant improvements in the detection of the fruit clusters. However, their evaluation focused on the viewpoint sampling strategy and did not clearly identify the advantage of adding an attention mechanism. Hence, there is no prior work that clearly studied the importance of an attention mechanism for active vision.

To the best of our knowledge, there are almost no prior work that evaluated the performance of active vision in real-world agro-food environments, such as a tomato greenhouse. They were extensively evaluated in simulation (Zaenker et al., 2021), with artificial plants (Lehnert et al., 2019), or small potted plants in a lab setup (Gibbs et al., 2018, 2019). Recently, Yi et al. (2024) tested an active-vision algorithm in a real-world environment for grape harvesting, but they did not study the impact of adding an attention mechanism. Hence, the advantage of an attention mechanism for active-vision algorithms was never validated under real-world conditions, such as natural variation and occlusion, natural illumination, sensor noise, and uncertainty in camera poses, which could strongly influence their performance.

This paper aimed to address these gaps by answering the questions: What is the improvement in performance of active vision, in terms of accuracy and speed, when an attention mechanism is added to it? Is this improvement in performance, if any, also valid under real-world conditions? By answering these questions, we can clearly identify the need for an attention mechanism for fast and accurate perception of targeted plant parts in complex greenhouse environments. A simple attention mechanism that could easily be combined with existing active-vision algorithms was first developed. This enabled the active-vision algorithm to focus only on the relevant plant parts when performing a greenhouse task. This strategy was then evaluated on the task of 3D plant reconstruction with three levels of attention – (i) whole plant, (ii) main stem, and (iii) leaf nodes. These levels of attention were chosen because they reflect the regions-of-interest (ROIs) required to perform three different tasks in a greenhouse, namely plant growth monitoring, visual search of plant nodes for harvesting and de-leafing, and pose estimation of leaf nodes for de-leafing respectively. The attention-driven active-vision strategy was evaluated on each attention level and compared to predefined and random multi-view perception strategies. The evaluation was done with both simulation and real-world data. In the simulated environment, we used 3D mesh models of tomato plants with varying growth stages and structural complexity. The real-world tomato plants added further complexity in terms of natural variation and occlusion, sensor noise, and uncertainty in camera pose. Using these setups, systematic, repeatable, and statistically significant experiments were performed to study the advantage of using an attention mechanism for active vision. Also, to the best of our knowledge, this work was the first to test the performance of active vision using a robotic setup in a real-world greenhouse environment.

## 2. Attention-driven next-best-view planning

### 2.1. Problem definition

The most common formulation of active vision is the next-best-view (NBV) planning problem (Bircher et al., 2018; Delmerico et al., 2018; Devrim Kaba et al., 2017; Lehnert et al., 2019; Mendoza et al., 2020; Peralta et al., 2020). Based on the NBV planning problem, our goal was to determine the next camera viewpoint that will perceive the most novel information about an object. By maximising the amount of novel information per viewpoint, the object can be completely reconstructed with the least number of viewpoints.

An OctoMap (Hornung et al., 2013) was used to represent the 3D space observed by the robot. An OctoMap converted measurements from a depth sensor, typically in the form of a point cloud, into a probabilistic occupancy map. The map divided the space into small grid cells known as *voxels* and encoded them as an octree (Meagher, 1982) for efficient computation. Each voxel was marked as free or occupied based on the probability of the voxel being occupied ($p_o$), as illustrated in Fig. 2. Voxels with an occupancy probability of zero ($p_o = 0$) and one ($p_o = 1$) were considered as free and occupied respectively with absolute certainty. The values in between implied some uncertainty regarding the occupancy, with maximum uncertainty at $p_o = 0.5$. The volumetric information $I_v$ expected to be gained by observing a voxel $x$ was then defined as its Shannon entropy (Arbel & Ferrie, 2001):

$$I_v(x) = -p_o(x)\log_2(p_o(x)) - (1 - p_o(x))\log_2(1 - p_o(x)), \quad (1)$$

where $p_o(x)$ was the probability of voxel $x$ being occupied. The volumetric information gain was also maximum when $p_o(x) = 0.5$, that is, when we were most uncertain about the occupancy of the voxel.

The view selection process in NBV planning was guided by the metric of *expected information gain*. In this context, information gain for a viewpoint ξ was defined as the cumulative volumetric information expected to be gained from all voxels visible from the viewpoint. Hence, information gain $I_v(\xi)$ for a viewpoint ξ was given by,

$$I_v(\xi) = \sum_{x \in \mathscr{X}_\xi} I_v(x), \quad (2)$$

where $\mathscr{X}_\xi$ was a set of all voxels that were expected to be visible from viewpoint $\xi$. $\mathscr{X}_\xi$ was calculated by an operation called *ray tracing*, in which a set of rays were cast from viewpoint $\xi$ evenly across its view frustum, as illustrated in Fig. 3. Each ray traverses until it encountered an occupied voxel or reached a maximum distance. All voxels that the rays intersected, both free and occupied, were expected to be visible if the camera was moved to viewpoint $\xi$ and hence were added to the set $\mathscr{X}_\xi$. Then, the cumulative volumetric information gain over $\mathscr{X}_\xi$ gave the expected information gain of viewpoint $\xi$. Intuitively, a viewpoint that observed the largest number of highly uncertain voxels generated a larger gain in information. The objective of the NBV planning problem was to determine the next-best viewpoint that maximised the information gain.

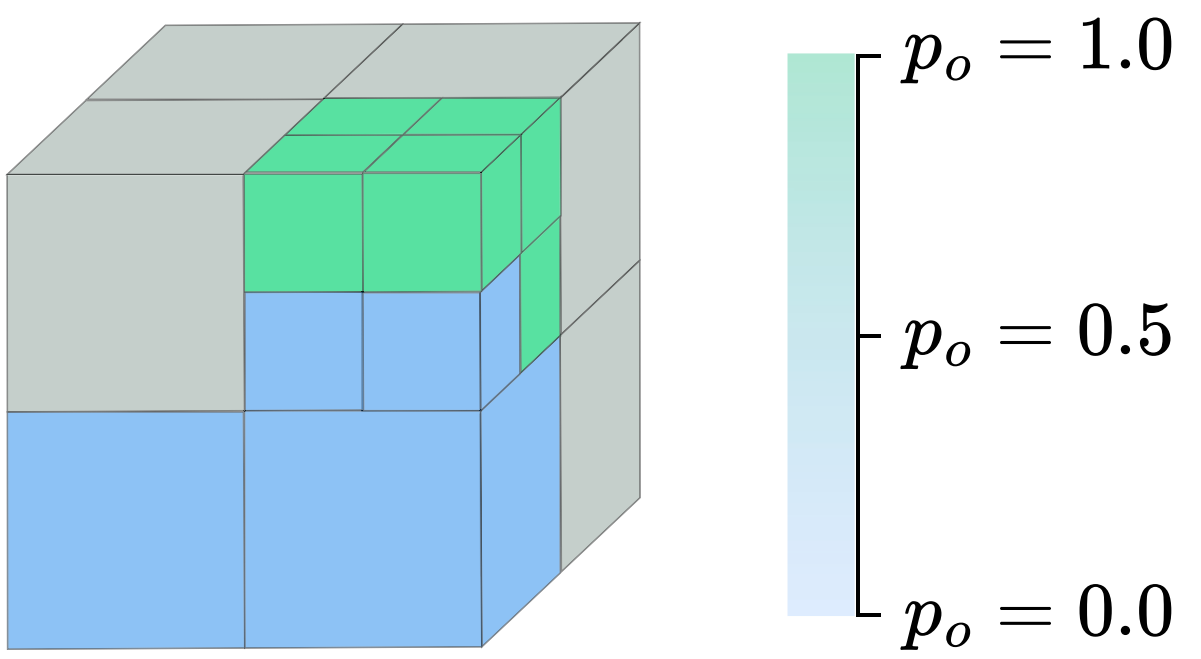

**Fig. 2.** Octree representation illustrating occupied (green) and free (blue) voxels with the probability of being occupied ($p_o$) equal to one and zero respectively. The unknown or unobserved regions (grey) are not part of the octree. (For interpretation of the references to colour in this figure legend, the reader is referred to the Web version of this article.)

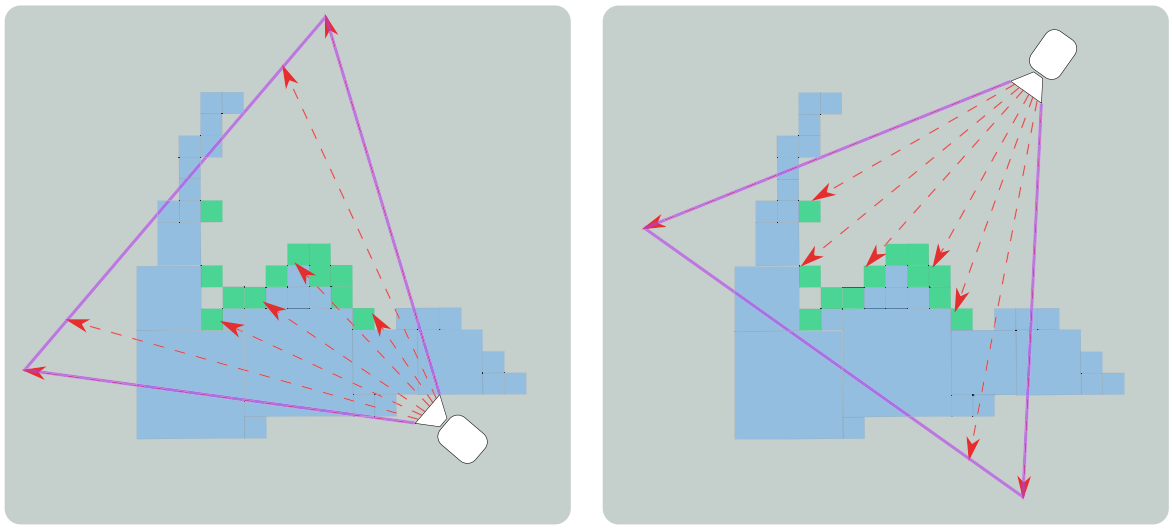

**Fig. 3.** Illustration of the ray-tracing operation from two candidate viewpoints (camera in white). The octree consists of occupied (green) and free (blue) voxels. The unknown or unobserved regions (grey) are also shown. Rays (red) are cast from a viewpoint evenly across its view frustum (top-view shown in purple) to predict which voxels will be visible from the candidate viewpoints. In the right situation, the rays pass through more unknown regions, resulting in a higher expected information gain. (For interpretation of the references to colour in this figure legend, the reader is referred to the Web version of this article.)

### 2.2. Next-best-view planning

The implementation of Isler et al. (2016) and Delmerico et al. (2018) for volumetric NBV planning was followed. The general pipeline, as illustrated in Fig. 4, had three major stages that were executed iteratively: (I) extracting depth information from the current viewpoint, (II) randomly sampling a set of candidates for the next viewpoint, and (III) estimating the expected information gain for each candidate viewpoint and determining the next-best viewpoint.

**Stage I.** An RGB-D camera was used to extract the depth information from the scene, which was converted to a point cloud. The point cloud was then inserted into an OctoMap, which divided the space into occupied and free voxels, as elaborated in Section 2.1. The resolution of the OctoMap was set to 0.003 m, which was fine enough to reconstruct details of small plant parts such as leaf nodes. The rest of the OctoMap parameters were set to their default values as defined by Hornung et al. (2013). Each point in the point cloud was assumed to correspond to an object surface and hence added as an occupied voxel with $p_o(x|z_t) = 0.7$, while the space along the line-of-sight between the camera origin and the point were added as free voxels with $p_o(x|z_t) = 0.4$. Here, the measurement $z_t$ referred to whether the voxel was free or occupied. When a voxel already existed in the space where a new measurement was made, the occupancy probability of the voxel was updated using the formula,

$$L(x \mid z_{1:t}) = L(x \mid z_{1:t-1}) + L(x \mid z_t) \quad (3)$$

$$\text{with} \quad L(x \mid z_t) = \log\left[\frac{p_o(x \mid z_t)}{1 - p_o(x \mid z_t)}\right], \quad (4)$$

where $L(x|z_t)$ is the log-odds notation of the occupancy probability $p_o(x|z_t)$ of voxel $x$. The update rule followed the Markov assumption, that is, the current occupancy probability of voxel $x$ given all depth measurements $z_{1:t}$ depended on the current measurement $L(x|z_t)$ and the previous estimate $L(x|z_{t-1})$. A more detailed explanation about the update rule is provided by Hornung et al. (2013). By applying this update rule iteratively, the OctoMap kept track of occupied and free voxels across multiple viewpoints. For updating the OctoMap, we only used the camera data acquired while the robot was stationary to prevent computational overhead and motion blur and improve synchronisation between the camera data and the camera pose.

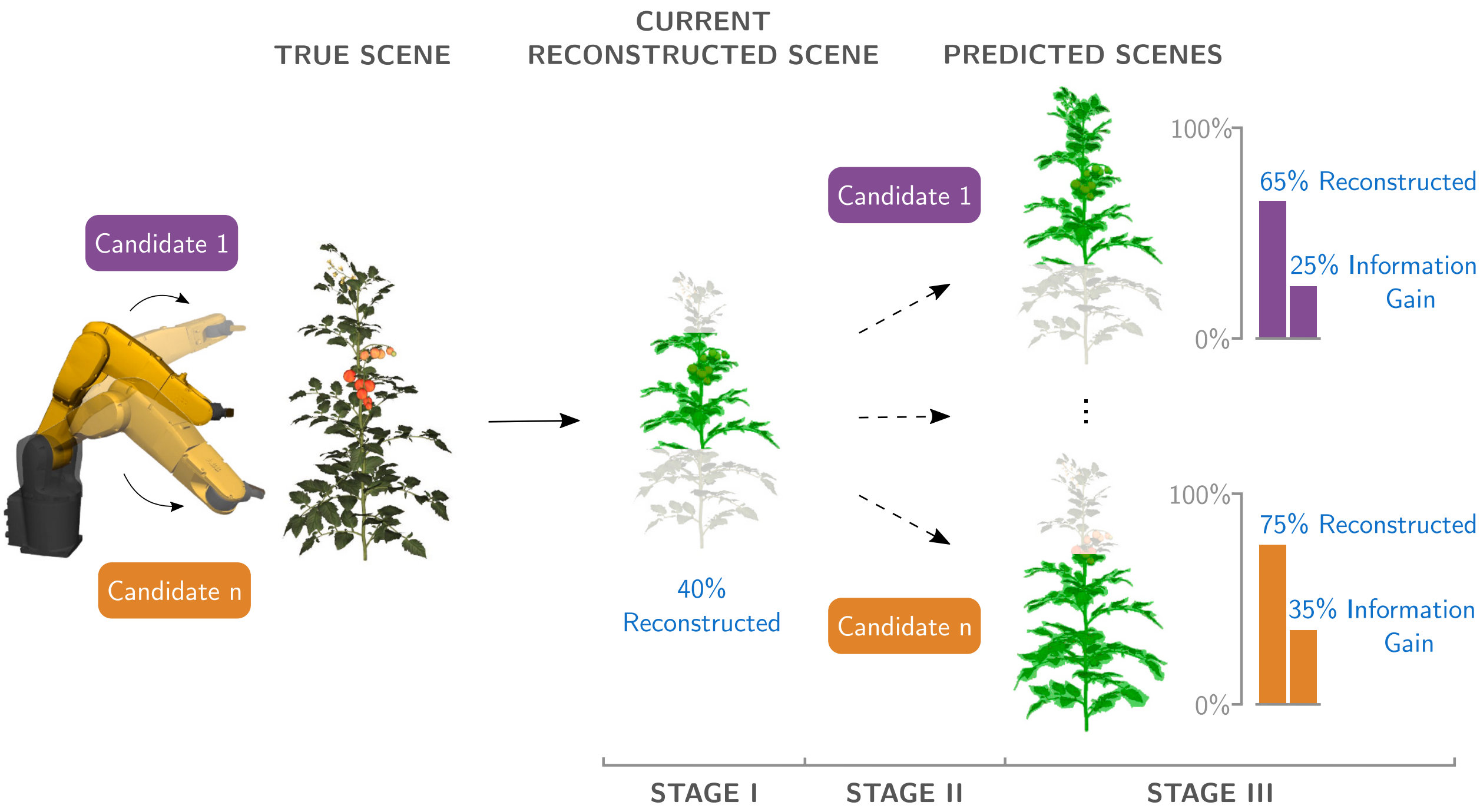

**Fig. 4.** Generic pipeline of a volumetric next-best-view planning algorithm with the three major stages – (I) extraction of information from current viewpoint, (II) sampling of a set of candidates for the next viewpoint, and (III) estimation of information gain and determining the next-best viewpoint.

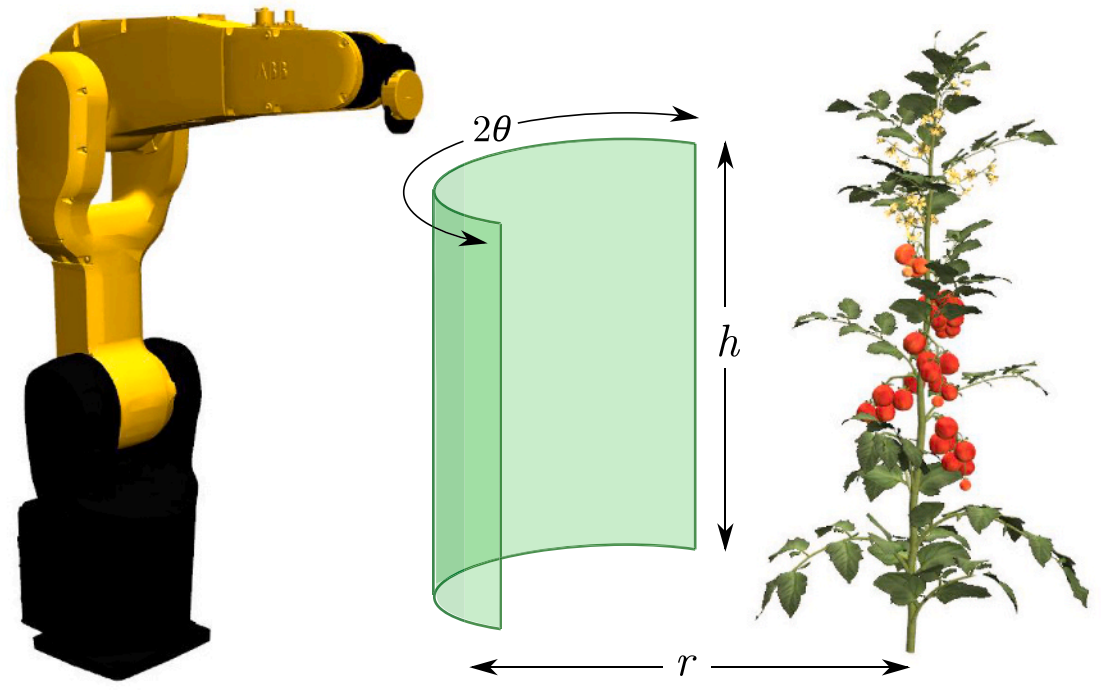

**Fig. 5.** Schematic diagram of the experimental setup. The robot constitutes of ABB's IRB 1200 with an Intel Realsense L515 camera. The camera viewpoints are constrained to a cylindrical sector (shown in green), with height h, radius r, and sector angle 2θ. The tomato plant is placed at the center of the cylinder. (For interpretation of the references to colour in this figure legend, the reader is referred to the Web version of this article.)

**Stage II.** A set of candidate viewpoints $\mathscr{V}$ was sampled at each step. These are potential positions where the camera can be moved next. We constrained the candidates to a cylindrical sector as shown in Fig. 5, so that the camera was always oriented towards the plant. In the absence of such constraint, an extensively large number of candidates need to be sampled so that at least a few of them have the plant in view. Furthermore, a pseudo-random strategy was used for sampling, in which the surface of the cylindrical sector was divided into a 3-by-3 grid of equal sizes and 3 viewpoints were randomly sampled from each of these grids. This strategy ensured the sampling of viewpoints across different regions of the cylindrical-sector surface. Moreover, the 3-by-3 grid split the viewing space such that each row of the grid could observe the top, middle, and bottom parts of the plant, while each column could observe the left, middle, and right parts of the plant. We found the pseudo-random sampling necessary so that the NBV planner can consider a diverse set of viewpoints at each step, before picking the next-best viewpoint. Otherwise, in situations where all candidates are sampled from the same region on the cylindrical sector, the NBV planner will be forced to pick a viewpoint despite all of them being poor candidates. Although the next iteration of view planning might generate better samples, the planner will end up with one extra viewpoint that could have been avoided.

**Stage III.** The expected information gain was estimated for all the candidate viewpoints $\xi \in \mathscr{V}$. For each candidate viewpoint $\xi$, the set $\mathscr{X}_\xi$ of potentially visible voxels were predicted with ray tracing and the expected information gain was estimated using Equation (2), that is, by summing the volumetric information gain $I_v$ for each predicted voxel in $\mathscr{X}_\xi$. This process of ray tracing and estimating the information gain was repeated for all the candidate viewpoints and the viewpoint with the highest gain was selected as the next-best viewpoint, that is:

$$\xi_{\text{best}} = \underset{\xi \in \mathscr{V}}{\arg\max} I_v(\xi). \tag{5}$$

In the ray-tracing operation, the maximum distance that a ray could travel before termination was set to 0.75 m. This range allowed the rays to traverse the complete width of the plant. All voxels beyond this range were ignored.

### 2.2.1. *Attention mechanism*

On top of the volumetric NBV algorithm, we added an attention mechanism to deal with targeted perception objectives. To focus attention on a certain region in space, we define a bounding box encompassing the region of interest (ROI). In particular, the bounding box was defined as a set of cuboids using 6 parameters – its centre point and dimensions of the sides. The attention mechanism was deliberately kept simple so that it could easily be integrated

**Algorithm 1** Attention-driven next-best-view planning – Iterative step

$G_{\text{best}} \leftarrow 0$ ▷ Initialise the best gain
$\xi_{best} \leftarrow \xi_0$ ▷ Initialise the best viewpoint
Randomly sample a set of candidate viewpoints $\mathscr{V}$
**for** viewpoint $\xi$ in set $\mathscr{V}$ **do**
  $G_v(\xi) \leftarrow 0$ ▷ Initialise gain for viewpoint
  Collect set $\mathscr{X}_\xi$ with ray tracing
  **for** voxel $x$ in set $\mathscr{X}_\xi$ **do**
    **if** voxel $x$ is in set $\mathscr{B}$ then
      $G_v(\xi) \leftarrow G_v(\xi) + I_v(x)$
    **end if**
  **end for**
  **if** $G_v(\xi) > G_{best}$ then
    $G_{best} \leftarrow G_v(\xi)$ ▷ Update the best gain
    $\xi_{best} \leftarrow \xi$ ▷ Update the best viewpoint
  **end if**
**end for**
return $\xi_{best}$ ▷ Next-best viewpoint and evaluated with existing active-vision algorithms. It was incorporated into the original NBV algorithm at Stage III. During the ray-tracing operation from viewpoint $\xi$, only the voxels in set $\mathscr{X}_\xi$ that were within the ROI were considered for the calculation of information gain $I_v(\xi)$. All voxels outside the ROI, irrespective of their state, were ignored. Hence, Equation (2) was modified as,

$$I_v(\xi) = \sum_{x \in (\mathscr{X}_\xi \cap \mathscr{B})} I_v(x), \tag{6}$$

where $\mathscr{B}$ is the set of voxels that were inside the ROI. The modified information gain $I_v(\xi)$ only considered voxels that were both visible from viewpoint $\xi$ and were inside the ROI, that is, $x \in (\mathscr{X}_\xi \cap \mathscr{B})$. The iterative step of our attention-driven NBV planner is summarised by Algorithm 1 and the parameters used are defined in Table 1.

The attention-driven NBV planner was subjected to two constraints – the cylindrical sector on whose surface candidate viewpoints could be sampled in Stage II and the bounding box constraints that were used to define the relevant parts of the plant in Stage III. Although both constraints incorporated a form of attention, they had distinct functions. The cylindrical-sector constraint ensured that the sampled candidate viewpoints always had the plant in view and were oriented towards the main stem, while the bounding box constraint was used by the NBV planner to prioritise relevant parts of the plant that had not been observed yet. In this paper, both constraints were defined using prior knowledge about the exact positions of the plant, main stem, and leaf nodes. We assumed that the exact positions were known since our focus was to study the advantage of adding an attention mechanism with ROIs and not on any specific method for finding and adding ROIs. However, in practice, the positions of these parts are not immediately available. There are some ways to automatically estimate the position of relevant plant parts using detection algorithms, as discussed in Section 5.3.

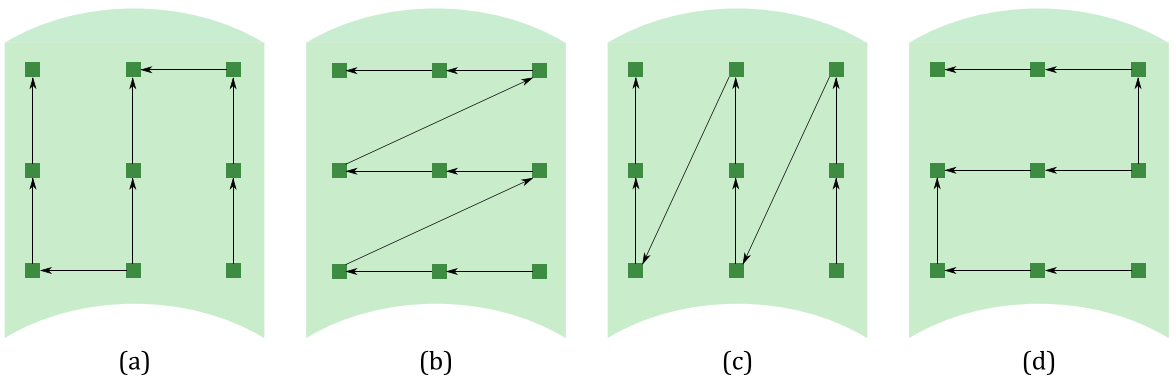

**Fig. 6.** Four different patterns (a–d) were used for the predefined planner. The green squares indicate the camera viewpoints. The predefined planner visited each camera viewpoint in the order illustrated by the arrows, starting from the bottom-right corner. (For interpretation of the references to colour in this figure legend, the reader is referred to the Web version of this article.)

## 3. Experiments

### *3.1. Baseline methods*

As baselines, we used predefined and random planning strategies. For the predefined planner, we chose a set of 9 viewpoints, each defined at the centre of the 3-by-3 grid on the surface of the cylindrical sector. The planner visited all the viewpoints in a predefined order and looped over them when more than 9 viewpoints were needed. The pattern of visiting each predefined viewpoint could be defined in multiple ways and depending on the scene, some patterns could have an undue advantage over others. So, we used four different patterns, as shown in Fig. 6, and took the average reconstruction performance over all of them. These predefined patterns represented the view planning strategies that are typically used in similar robotic applications (Boogaard et al., 2020).

**Table 1**
Experimental parameters for simulation and real-world experiments.

| Parameter | | Simulation Values | Real-world Values |
|---|---|---|---|
| Center of the plant | Position[a] | **x:** 1.0 m **y:** 0.0 m **z:** 1.15 m | **x:** 0.0 m **y:** −0.6 m **z:** 0.9 m |
| | Initial orientation[a] | **x:** 0.0 **y:** 0.0 **z:** 0.0 **w:** 1.0 | **x:** 0.0 **y:** 0.0 **z:** 0.0 **w:** 1.0 |
| Initial camera viewpoint | Position[a] | **x:** 0.646 m **y:** 0.353 m **z:** 1.383 m | **x:** 0.169 m **y:** −0.354 m **z:** 0.54 m |
| | Orientation[a] | **x:** 0.0 **y:** 0.0 **z:** −0.383 **w:** 0.924 | **x:** 0.213 **y:** 0.665 **z:** −0.683 **w:** −0.213 |
| Cylindrical sector | Height | **h:** 0.7 m | **h:** 0.7 m |
| | Radius | **r:** 0.3 m | **r:** 0.3 m |
| | Sector angle | **2θ:** 90° | **2θ:** 90° |
| Octomap[b] | Resolution | 0.003 m | 0.003 m |
| | Tree depth | 16 | 16 |
| | Max range | 0.75 m | 0.75 m |
| | Raycast range | 0.75 m | 0.75 m |
| | Occupancy threshold | 0.5 | 0.5 |
| | Clamping threshold | **min:** 0.12, **max:** 0.97 | **min:** 0.12, **max:** 0.97 |
| Size of bounding box | Whole plant | **x:** 0.3 m, **y:** 0.3 m **z:** 0.7 m | **x:** 0.4 m, **y:** 0.4 m **z:** 1.2 m |
| | Main stem | **x:** 0.05 m, **y:** 0.05 m, **z:** 0.7 m | – |
| | Leaf nodes | **x:** 0.03 m, **y:** 0.03 m, **z:** 0.05 m | **x:** 0.06 m, **y:** 0.06 m, **z:** 0.06 m |
| Position of bounding box center | Whole plant | **x:** 1.0 m, **y:**0.0 m **z:** 1.15 m | **x:** 0.0 m, **y:** −0.6 m **z:** 0.9 m |
| | Main stem | **x:** 1.0 m, **y:**0.0 m **z:** 1.15 m | – |
| | Leaf nodes | Varied based on node location | Varied based on node location |
| Planners | # of view candidates | 27 | 30 |
| | Maximum views | 10 | 10 |

[a] Values are with respect to a global world frame. Orientations are given as quaternions.
[b] Please refer to Hornung et al. (2013) for details regarding the OctoMap parameters.

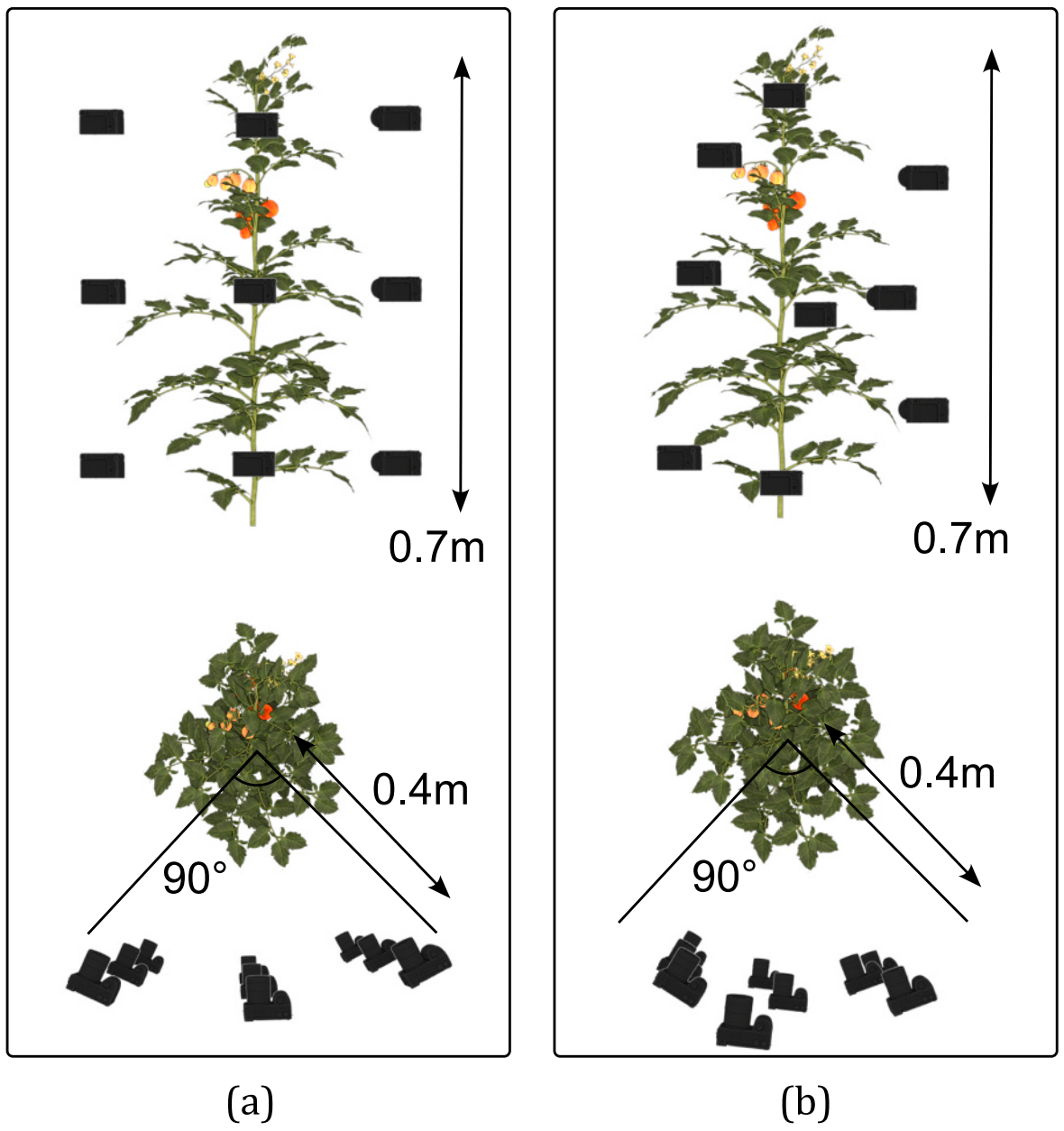

**Fig. 7.** An example of front view and top view of some candidate viewpoints sampled on the cylindrical surface by (a) the predefined planner and (b) the random planner. The black camera icons represent the candidate viewpoints. The cylindrical surface on which the candidates were sampled has height h = 0.7 m, radius r = 0.4 m, and a sector angle 2θ = 90°.

For the random planner, one viewpoint was randomly chosen from the same set of candidate viewpoints $\mathscr{V}$ used for the NBV planners (see Section 2.2, Stage II). At each step, a new set of candidate viewpoints was sampled and one viewpoint among them was randomly chosen by the random planner.

To ensure that the performances of the planners were comparable, all of them started from the same initial viewpoint $\xi_0$. The choice of the initial viewpoint was arbitrary and did not influence the performance of the planners. Also, all planners were constrained to the same cylindrical sector for view selection. Fig. 7 illustrates some viewpoints sampled by the predefined and random planners.

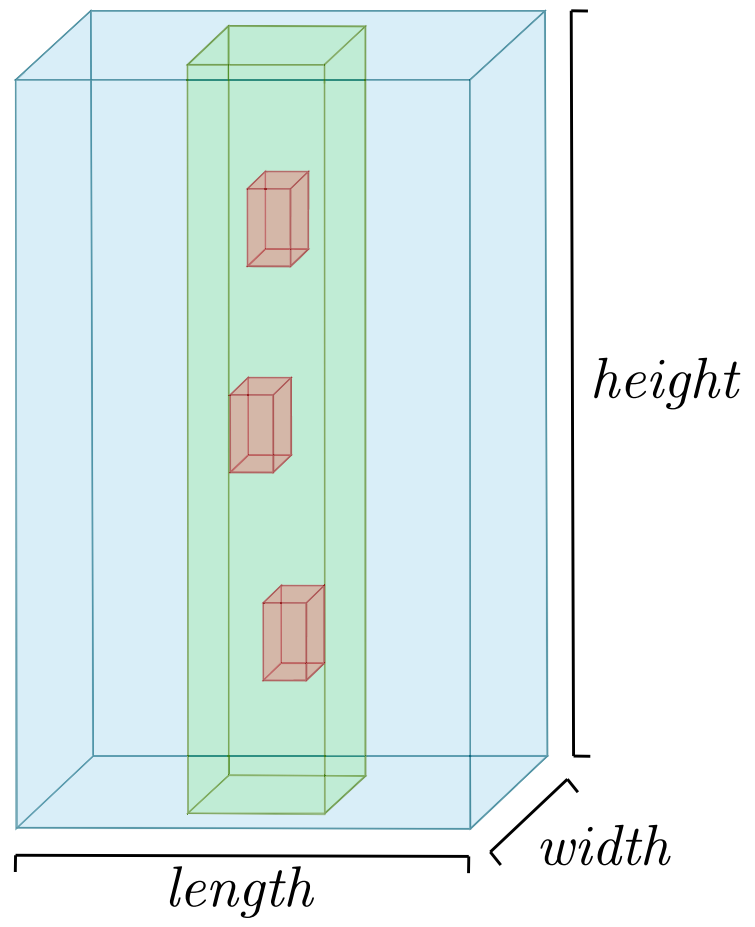

**Fig. 8.** An example of bounding boxes defining the regions of interest for different levels of attention – whole plant (blue), main stem (green) and leaf nodes (red). The exact positions and sizes of the bounding boxes are given in Table 1. (For interpretation of the references to colour in this figure legend, the reader is referred to the Web version of this article.)

### 3.2. Simulation experiments

We performed simulation experiments to evaluate the NBV planner on perception tasks that required different levels of attention. To test this, we considered the task of 3D plant reconstruction and applied three levels of attention, ranging from wide to narrow – (i) whole plant, (ii) main stem and (iii) leaf nodes, as illustrated in Fig. 8. For the remainder of the paper, we will refer to these NBV planners as NBV-whole-plant, NBV-main-stem, and NBV-leaf-nodes. The bounding box defining the ROI around the whole plant had a length and breadth of 0.3 m and a height of 0.7 m. The bounding box around the main stem had the same height but a narrower length and breadth of 0.05 m, which left out most parts of the leaves. The 3 bounding boxes around the leaf nodes were more constrained compared to the main stem bounding box, with a length and breadth of 0.03 m and a height of 0.05 m. The leaf nodes were chosen at different positions across the plant to ensure that the reconstruction performances of the view planners were invariant to the location of the leaf nodes. The sizes for the bounding boxes were chosen such that they encapsulated the whole plant, the main stem, and the leaf nodes.

#### 3.2.1. Simulation setup

The simulation setup consisted of a 6DoF robotic arm (ABB's IRB 1200) with an RGB-D camera (Intel Realsense L515) attached to it, as illustrated by Fig. 5. We placed a tomato plant directly in front of the robot at a distance of 1.0 m. The robotic arm was allowed to move the camera to different positions on a cylindrical sector centred around the stem of the plant with a radius of 0.4 m, height of 0.7 m, and a sector angle of 90°. The parameters of the cylinder were chosen such that it was wide enough to observe the complete plant and ensure that all the viewpoints on the cylinder were reachable by the robotic arm without collisions. Hence, the robot could observe the tomato plant from multiple viewpoints and reconstruct it. The robot was allowed a maximum of 10 viewpoints for the reconstruction task.

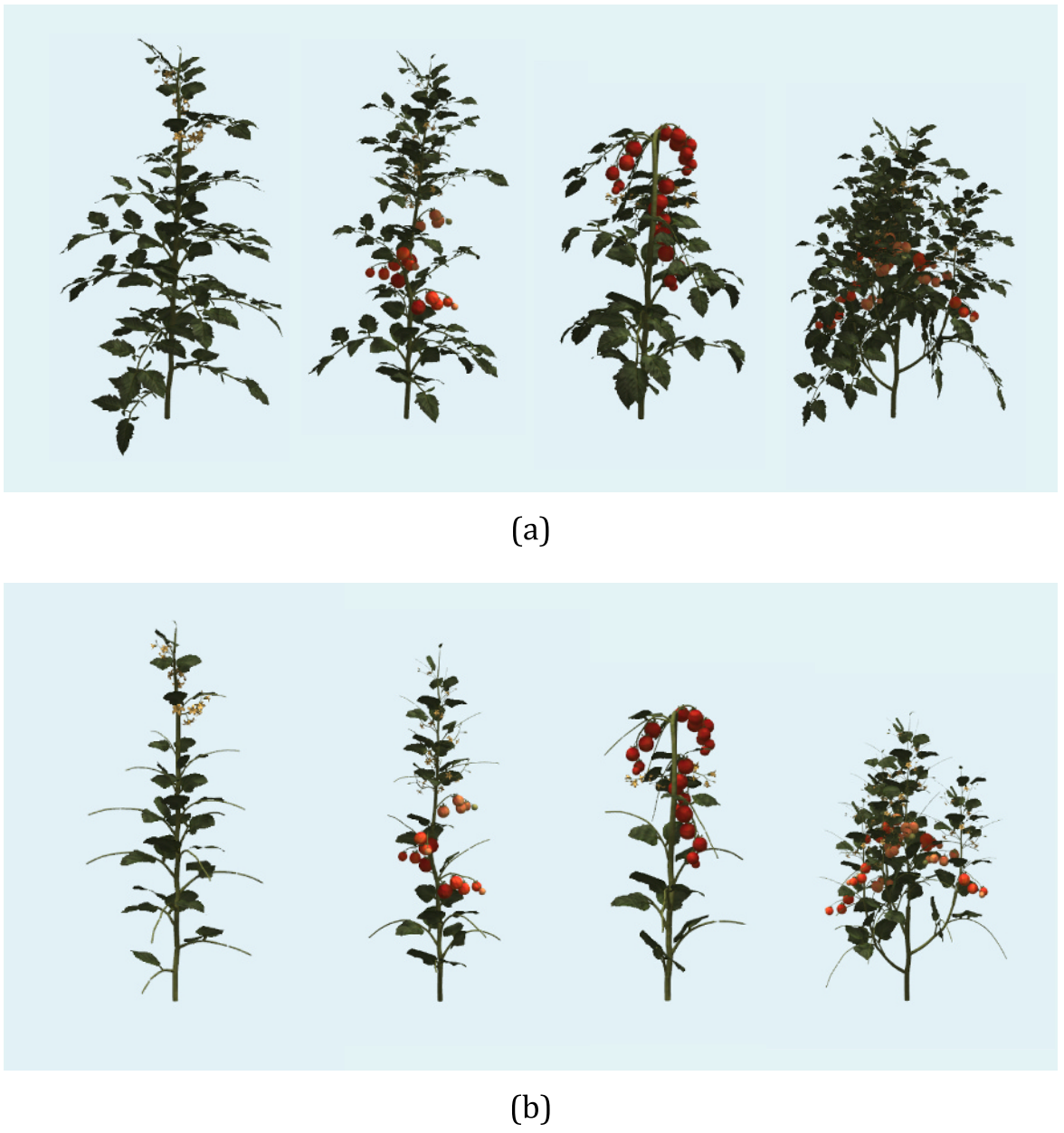

**Fig. 9.** Examples of (a) the original plant models and (b) the same plant models with some leaflets removed that were used in simulation.

**Table 2**
Details about the geometric and structural complexity of the tomato plants used in simulation experiments.

| Plant | Number of vertices | Number of leaf nodes | Number of fruit nodes | Plant | Number of vertices | Number of leaf nodes | Number of fruit nodes |
|---|---|---|---|---|---|---|---|
| **Plant 1** | 32,652 | 17 | 0 | **Plant 6** | 78,434 | 15 | 4 |
| **Plant 2** | 50,146 | 14 | 0 | **Plant 7** | 104,602 | 12 | 5 |
| **Plant 3** | 60,016 | 16 | 0 | **Plant 8** | 60,988 | 12 | 4 |
| **Plant 4** | 67,704 | 15 | 3 | **Plant 9** | 45,124 | 13 | 4 |
| **Plant 5** | 42,888 | 21 | 2 | **Plant 10** | 230,896 | 59 | 11 |

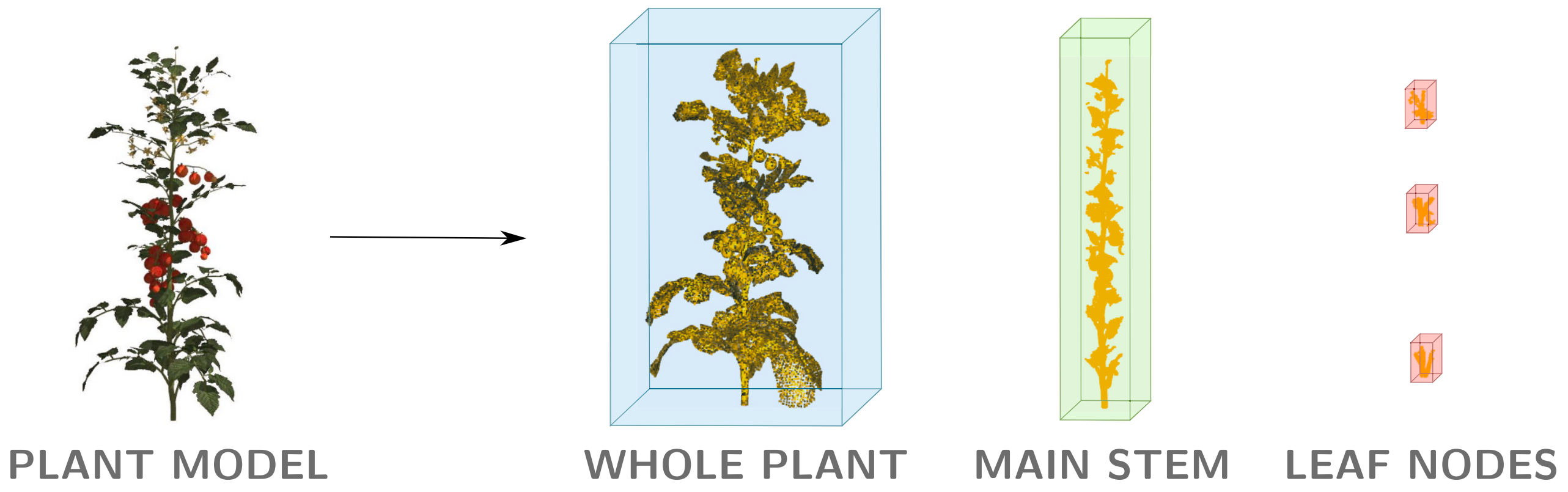

**Fig. 10.** The plant models were trimmed according to the three levels of attention – whole plant, main stem, and leaf nodes. The trimmed models were used as ground truth to evaluate the 3D reconstruction performance of the view planners.

The experiments were conducted using a simulated environment in *Gazebo*.[1] Ten 3D mesh models of tomato plants[2] were used to conduct the experiments. The models had tomato plants with different growth stages and hence they showed variation in height, structure, and number of tomato trusses and leaf nodes, as shown in Fig. 9a and Table 2. For each plant, the experiments were repeated with 12 different orientations of the plant, with a 30° difference between each orientation. Overall, a total of 120 trials were conducted using the 10 plant models.

### 3.2.2. *Influence of experimental and model parameters*

The experimental conditions and parameters (defined in Table 1) might have a profound impact on the performance of the planners. Hence, we studied the influence of changes in the experimental conditions and parameters on the reconstruction of leaf nodes. In particular, the effect of the amount of occlusion, the number of candidate viewpoints at each step, and the resolution of the OctoMap used for reconstruction were studied. Such an analysis would help us understand the conditions under which an attention mechanism is favourable for targeted perception objectives.

**Amount of occlusion**. The amount of occlusion was reduced by removing some leaflets of the plant models, as shown in Fig. 9b. In particular, half the leaflets on each leaf were removed. With less occlusion, we expected that the NBV planner would be able to reconstruct the plant and plant parts in less viewpoints and more accurately.

**Number of candidate viewpoints**. The number of candidate viewpoints that the NBV algorithm could consider at each step was varied to 9 and 45, from the original size of 27. With respect to the pseudo-random sampling strategy, this corresponded to sampling 1 and 5 viewpoints per grid, from the original of 3 viewpoints per grid. The number of candidates were chosen such that there was at least one candidate per grid for pseudo-random sampling, which ensured that the candidates were sampled from different regions of the cylinder. On the other hand, increasing the number of candidates would increase the computational cost. So, it was limited to a feasible number. When the number of candidate viewpoints are higher, we expected that the NBV planner would have a greater chance of finding a viewpoint with a large information gain.

**Resolution of reconstruction**. The voxel resolution of the OctoMap was reduced to 0.005 m and 0.007 m, from the original resolution of 0.003 m. The evaluation criteria for the reconstruction performance, detailed in Section 3.2.3, were adjusted according to the voxel resolution. The voxel resolutions were chosen such that the voxels could represent thin structures of the plant, such as leaves, while still being computationally feasible. So, we limited the resolutions between 0.003 m and 0.007 m. We expected that minor changes in the resolution of the OctoMap will not have much impact on the performance of the NBV planner.

### 3.2.3. *Evaluation metrics*

We compared the performance of the attention-driven NBV planners on the task of 3D plant reconstruction with three levels of attention. We evaluated the planners on two criteria – (i) the accuracy and (ii) the speed of reconstruction. Accuracy was determined by evaluating the quality of reconstruction compared to the ground truth, that is, the original mesh model of the tomato plants used in simulation. For the sake of evaluation, both the ground-truth mesh model and the reconstructed OctoMap were converted to point clouds. The ground-truth mesh model was converted to a point cloud by uniformly sampling points on the mesh surface and then down sampling the points using a voxel-grid filter to a resolution of 0.003 m, that is, the original resolution of the OctoMap. To evaluate the reconstruction accuracy for the three different tasks (whole-plant, main-stem and leaf-node reconstruction), the point clouds were trimmed according to the dimensions of the bounding boxes, as illustrated in Fig. 10. Note that, since 3 regions of interest were defined for the leaf nodes, the performance of the planners on each individual node were averaged to get the overall performance.

We applied the following metrics to determine accuracy and speed of each planner:

**F1-score.** We evaluated the accuracy using the F1-score. The F1-score is defined as the harmonic mean between precision and recall. Precision refers to the correctness of the reconstruction and identifies the fraction of reconstructed points ($r \in R$) that lie within a certain distance $\rho$ to the ground truth. Recall refers to the completeness of the reconstruction and identifies the fraction of ground-truth points ($t \in T$) that lie within a certain distance $\rho$ to the reconstruction. We used a distance threshold equal to the voxel resolution, that is, $\rho = 0.003$ m to calculate precision and recall. This was the minimum value that we could use for the distance threshold, as it was limited by the voxel resolution. This threshold was also strict for evaluation, as it penalised the precision and recall even for a one-voxel error. The precision and recall are calculated by,

$$\text{precision} = \frac{\sum_{r \in R} \mathscr{I}\left\{\min_{t \in T} \|r - t\|_2 < \rho\right\}}{|R|}, \tag{7}$$

$$\text{recall} = \frac{\sum_{t \in T} \mathscr{I}\left\{\min_{r \in R} \|r - t\|_2 < \rho\right\}}{|T|} \tag{8}$$

where $\mathscr{I}$ is an indicator function which equals 1 if the predicate is true and 0 otherwise. Intuitively, the F1-score indicates the fraction of points that were reconstructed correctly. It takes a value between 0 and 1, where a value of 1 implies a complete and accurate reconstruction. The F1-score is given by,

$$F_1 = 2 \cdot \frac{\text{precision} \cdot \text{recall}}{\text{precision} + \text{recall}}. \tag{9}$$

[1] http://gazebosim.org/.
[2] https://www.cgtrader.com/3d-models/plant/other/xfrogplants-tomato.

**# Views.** We evaluated the speed of reconstruction as the number of views required by the planning algorithms to reach an accuracy threshold. We chose the accuracy threshold $\tau_a$ to be an F1-score of 0.8 and 0.9, that is, 80% and 90% completion of reconstruction. Due to the cylindrical-sector constraint, the robot could observe the plant only from one side. Hence, an F1-score of 1.0 was impossible to reach. We assumed that an F1-score between 80 and 90% is possible to achieve. The speed of reconstruction was then quantified as the number of views needed to reach the target accuracy of $F_1 > \tau_a$.

### 3.3. *Real-world experiments*

We performed experiments with data collected in a tomato greenhouse at Unifarm, Wageningen University and Research, The Netherlands, to test if the advantage of an attention mechanism for NBV planning extends to real-world conditions with natural variation and occlusion in tomato plants, natural illumination, sensor noise, and uncertainty in camera poses. We considered the task of 3D reconstruction of all the nodes along the main stem, as this information is needed for robotic harvesting and de-leafing tasks. We compared the performance of the NBV-leaf-nodes planner with the NBV-whole-plant, predefined, and random planners. The NBV-main-stem planner was excluded because the main stems of the real-world tomato plants were curved and far from vertical. So, an axis-aligned bounding box at the centre of the plant could not capture the ROI along the main stem. However, in future work, the method could be extended to deal with different stem shapes.

#### 3.3.1. *Real-world setup and data collection*

The real-world setup was similar to the simulation setup, with a 6DoF robotic arm (ABB's IRB 1200) and an RGB-D camera (Intel Realsense L515), as shown in Fig. 11a. The camera was mounted to the end-effector of the robotic arm and an eye-in-hand calibration was performed to determine the camera pose with respect to the robotic arm. The Robotic Operating System (ROS) (Quigley et al., 2009) was used as a middleware to run different processes such as data acquisition and motion planning. *MoveIt!* (Chitta et al., 2012), a motion planning package for ROS, was used for moving the camera to a desired viewpoint. The robot was mounted on a cart to easily move between rows of tomato plants in the greenhouse.

A dataset was collected in the greenhouse to record the colour and depth data of seven tomato plants. Each plant was observed from 100

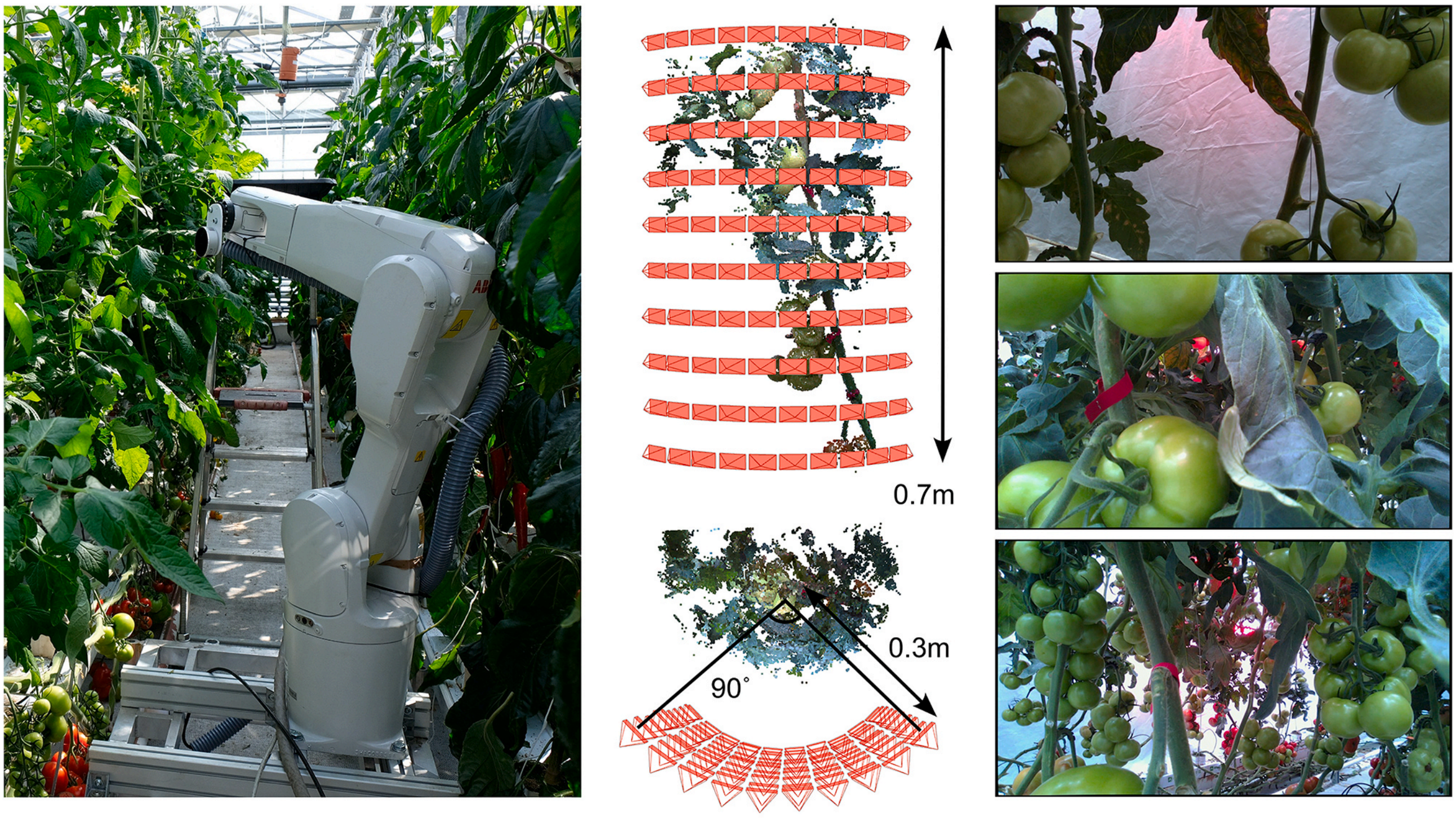

(a) (b) (c)

**Fig. 11.** (a) The real-world setup with a 6DoF ABB IRB 1200 robotic arm and a Realsense L515 RGBD camera. (b) Front and top views of the 100 viewpoints on a cylindrical sector of height h = 0.7 m, radius r = 0.3 m, and a sector angle 2θ = 90°, used for collecting data of tomato plants. (c) Example images of tomato plants collected by the robot, showing the level of occlusion.

different viewpoints on a cylindrical sector centred around the plant with a radius of 0.3 m, height of 0.7 m, and a sector angle of 90°. The 100 viewpoints were predefined along ten equally spaced rows and columns, as illustrated in Fig. 11b. For each viewpoint, a colour image of 960 × 540 pixels and the corresponding structured point cloud were recorded, along with the 6D camera pose with respect to the robot's base frame. In total, 25 nodes were observed across the seven plants. The plants in the greenhouse were mostly static due to the absence of wind and hence any small natural movements of the plants were not controlled. However, large movements caused when the robot accidently hit the plant were controlled by pausing the data collection process. Some examples of the captured images are shown in Fig. 11c.

For the real-world experiments, the dimensions of the ROIs were adjusted according to the size of real-world plants. The ROI for whole plant had a length and breadth of 0.4 m and a height of 1.2 m. The bounding boxes around the nodes had a length, breadth, and height of 0.06 m. The complete list of experimental parameters is given in Table 1. The objective of NBV planning was to determine the shortest sequence of viewpoints from the 100 candidates that could accurately reconstruct the plant nodes. The pre-recorded data allowed us to repeat experiments and compare different viewpoint planners. For each plant, the experiments were repeated 12 times leading to a total of 84 trials with seven real-world plants. The experiments were run on a laptop with a 2.60 GHz Intel Core i7-9750H CPU.

#### 3.3.2. *Ground truth and evaluation metrics*

Unlike the simulation experiments, the original mesh models of the tomato plants were obviously not available for the real-world experiments. Hence, we used COLMAP (Schonberger & Frahm, 2016), a structure-from-motion-based 3D reconstruction algorithm to reconstruct the tomato plants using the 100 pre-recorded colour images. The camera poses of the colour images were also provided to COLMAP, which ensured that the reconstruction was properly scaled and was with respect to the robot's base frame. The reconstructed points that were outside the ROI bounds defined for the whole plant (Table 1) were filtered out. The reconstructed point cloud from COLMAP was used as ground truth for evaluation (shown in Fig. 12a).

We evaluated the accuracy and speed of reconstruction of the planners for the real-world experiments. The steps for evaluation were identical to the simulation evaluation (Section 3.2.3), except for one detail. The distance threshold $\rho$ for computing the F1-score was increased to two-voxels distance from one, that is, $\rho = 0.006$. This was done to account for the fact that the COLMAP reconstruction was not the perfect ground truth and included some noise, which could give an error of one-voxel distance. With this modification, we compared the point cloud from the reconstructed OctoMap with the point cloud from COLMAP to quantify the accuracy of reconstruction (shown in Fig. 12b). As we were evaluating only the reconstruction of the plant nodes, we trimmed the COLMAP and the OctoMap point clouds to consider only the points within the ROIs, similar to the simulation evaluation.

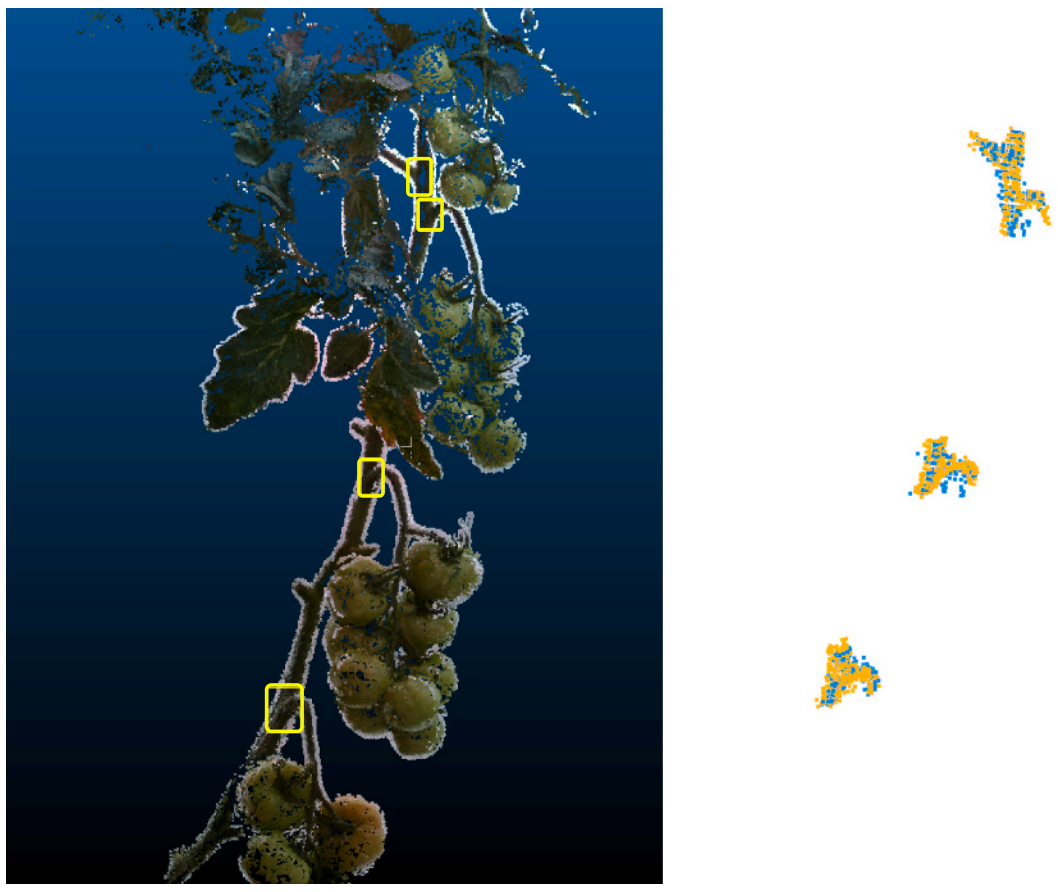

(a) (b)

**Fig. 12.** (a) Example of a reconstruction from COLMAP used as the ground truth for real-world experiments, along with the 3D node positions (marked in yellow). (b) The reconstructed point cloud from OctoMap (in blue) was compared with the ground-truth point cloud from COLMAP (in gold) using F1-score. (For interpretation of the references to colour in this figure legend, the reader is referred to the Web version of this article.)

## 4. Results

### 4.1. *Simulation results*

We report the performance of the NBV planners, with three levels of attention, on the reconstruction of tomato plants, with three levels of focus. The performances of the predefined and random planners are also reported as baselines. The results of reconstructing the whole plant are presented in Section 4.1.1, reconstructing the main stem in Section 4.1.2, and reconstructing the leaf nodes in Section 4.1.3. We also present the influence of the experimental parameters on the reconstruction of leaf nodes in Section 4.1.4. In all these experiments, the reconstruction performances of the planners varied across the 10 plant models and 12 starting orientations of the plants due to different levels of occlusions. We presented this variation as a 95% confidence interval over the 120 experiments conducted for each planner. We also performed a Kruskal-Wallis H Test to determine the statistical significance of our method. We illustrated the results of this test using asterisks in Fig. 13, which indicate the viewpoints for which the best-performing planner was statistically significant than all the other planners with a p-value less than 0.05.

#### 4.1.1. *Reconstruction of the whole plant*

Fig. 13a shows that for the task of reconstructing the whole plant, the NBV-whole-plant and NBV-main-stem planners performed the best. They both attained a higher F1-score in the initial viewpoints compared to the other planners. Both planners reached the accuracy threshold $\tau_a = 80\%$ in 3 viewpoints, 1 view faster than the predefined and random planners. The planners showed no significant differences in reaching $\tau_a = 90\%$. After 3 views, they reconstructed the whole plant 9.7% and 5.3% more than the predefined and random planners respectively in terms of their F1-score. After 10 views, all planners performed equally well (expect NBV-leaf-nodes), which implies that the predefined and random planners could also attain high reconstruction accuracy when more viewpoints are allowed. For our analysis, we considered the planner that attains high accuracy in the least number of viewpoints to be better. The performance of the planners saturates around an F1-score of 0.95 because the plants were observed only from one side. There were portions of the plants, particularly on the back side, which were not visible to the robot from any reachable viewpoint.

Within the NBV planners, there was no significant difference when the focus of attention was on the whole plant or the main stem. There might be two reasons for this indifference. First, the cylindrical-sector constraint that we defined in Section 3.2.1 forced the camera to always face the main stem, which prevented the NBV-whole-plant planner from going to viewpoints that were centred around other parts of the plant. Second, the field-of-view of the camera was wide enough to view the whole plant along the horizontal axis, which provided the NBV-main-stem planner the relevant information to reconstruct the whole plant. Due to these two aspects, the views sampled by both the NBV-whole-plant and NBV-main-stem planners contained similar information and led to similar performances. On the other hand, the NBV-leaf-nodes planner naturally performed the worst on this task, since it was directing attention to only small portions of the plant.

#### 4.1.2. *Reconstruction of the main stem*

For the task of reconstructing the main stem, the results were very similar to the reconstruction of the whole plant. The NBV-whole-plant and NBV-main-stem planners performed the best. The F1-scores of the reconstructions are shown in Fig. 13b. Both planners reached $\tau_a = 80\%$ in 4 viewpoints, 1 view faster than the predefined and random planners. The planners showed no significant differences in reaching $\tau_a = 90\%$. After 3 views, they reconstructed the main stem 14.2% and 7.9% more than the predefined and random planners respectively.

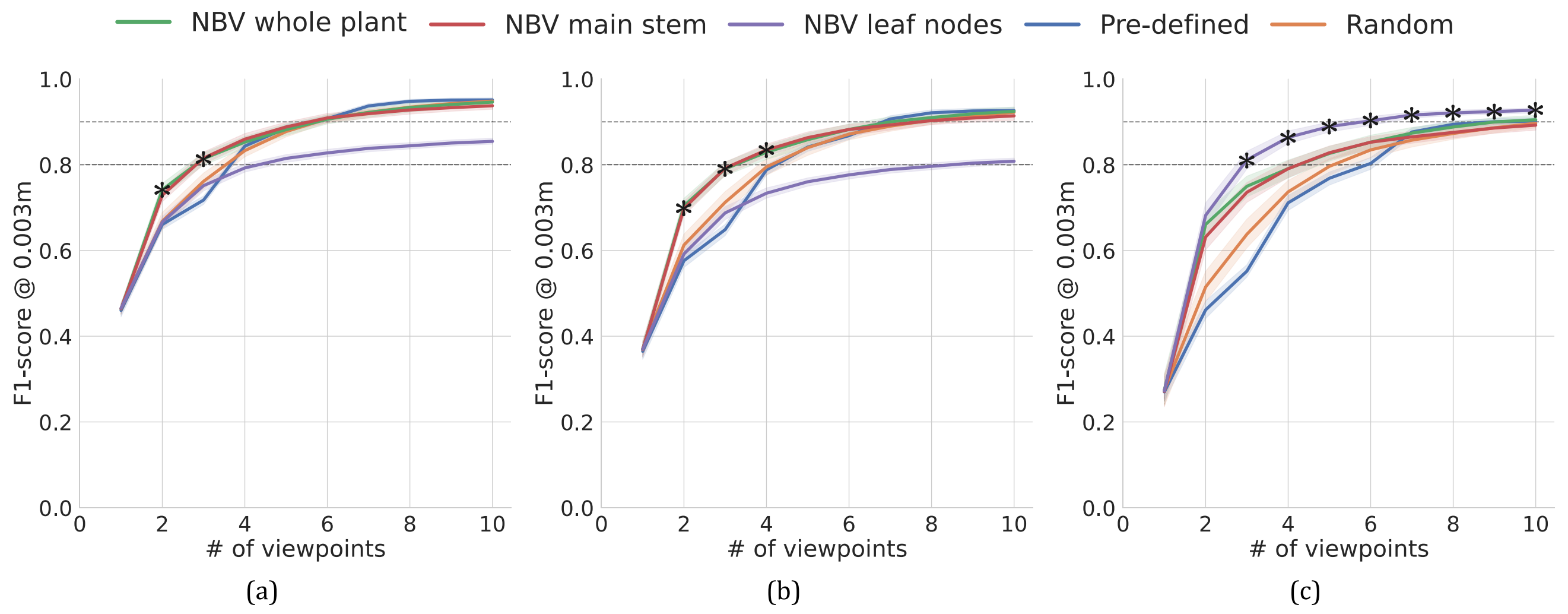

**Fig. 13.** Performance of the planning algorithms given by F1-score on the reconstruction of the (a) whole plant, (b) main stem, and (c) the leaf nodes. The shaded regions represent the 95% confidence interval of the mean over 120 experiments. The asterisks (*) denote the viewpoints for which the best-performing planner was significantly better than all the other planners with a p-value less than 0.05. Note that for plots (a) and (b), both the NBV-whole-plant and the NBV-main-stem planners were the best-performing planners, with no significant difference between them.

Again, we noticed no significant difference when the focus of attention was on the whole plant or the main stem, and the NBV-leaf-nodes planner performed the worst, as noted in Section 4.1.1. Compared to the performance on the reconstruction of whole plant, the NBV-whole-plant and NBV-main-stem planners took one extra viewpoint to reach $\tau_a = 80\%$. This is expected since the reconstruction of the main stem is a more targeted perception objective and would require more viewpoints to achieve the same completion rate of 80% in the presence of occlusions.

### 4.1.3. Reconstruction of leaf nodes

For the task of reconstructing the leaf nodes, we observed that the NBV-leaf-nodes planner significantly outperformed the other planners. The F1-scores of the reconstructions are shown in Fig. 13c. The NBV-leaf-nodes planner reached $\tau_a = 80\%$ in 3 viewpoints, 3 views faster than the predefined and random planners, and 2 views faster than NBV-whole-plant and NBV-main-stem planners. Also, it reached $\tau_a = 90\%$ in 6 viewpoints, at least 3 views faster than all the other planners. After 3 views, it reconstructed the leaf nodes 6.8%, 25.9%, and 17.3% more than the other NBV planners, predefined planner, and the random planner respectively.

The NBV-whole-plant and NBV-main-stem planners reached $\tau_a = 80\%$ in 5 viewpoints. After 5 viewpoints, the performances of the NBV-whole-plant and NBV-main-stem planners were not significantly different from the random planner, while the NBV-leaf-nodes planner significantly outperformed all of them. These results clearly indicate the significance of providing appropriate regions of interest for the attention mechanism. For the task of reconstructing the leaf nodes, focusing attention on the whole plant or the main stem does not work well, whereas appropriately focusing attention on the leaf nodes significantly improves the performance of the NBV planner. Hence, finding the appropriate regions of interest according to the task-at-hand is important.

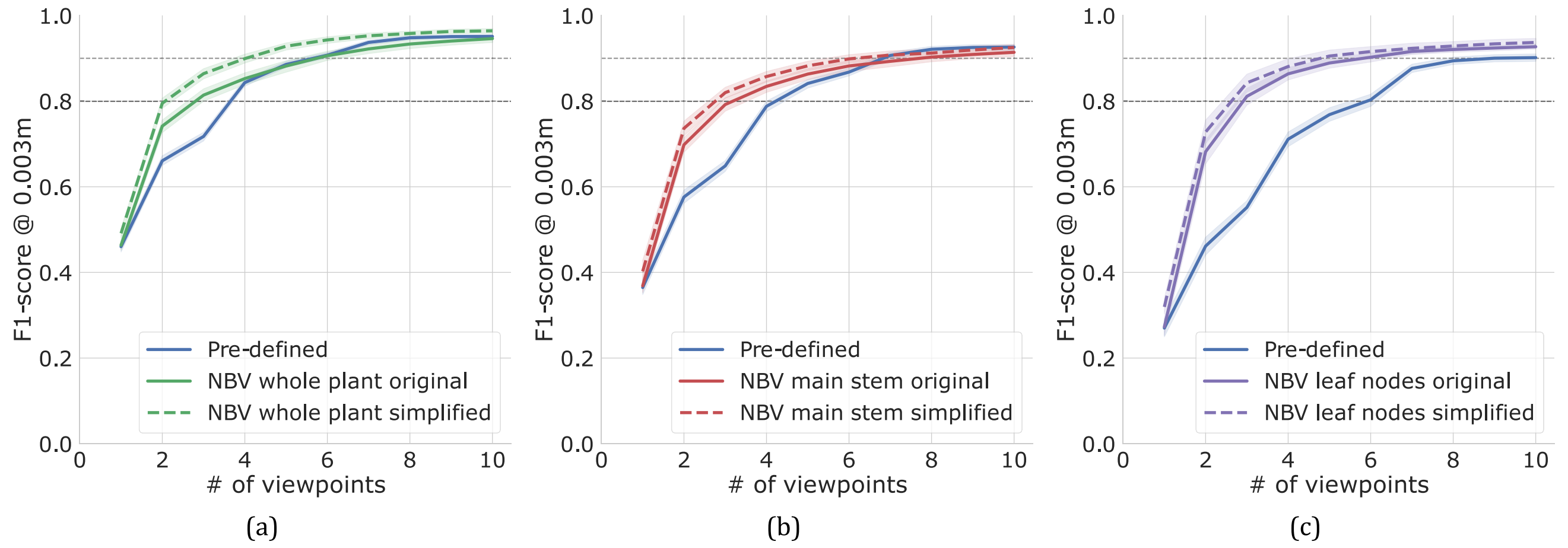

**Fig. 14.** Sensitivity of the performance of attention-driven NBV planner to changes in the amount of occlusion in the plants to be reconstructed. The plots show the comparison of F1-scores with the original plant models and simplified plant models with lower amount of occlusion (denoted as simplified), i.e., with some leaflets removed, for the reconstruction of (a) the whole plant, (b) the main stem, and (c) the leaf nodes. The shaded regions represent the 95% confidence interval of the mean.

#### 4.1.4. Influence of experimental and model parameters

We report the influence of experimental parameters on the performance of NBV planners on the reconstruction of leaf nodes.

**Amount of occlusion**. From Fig. 14, we observed that when the amount of occlusion was reduced, all the NBV planners performed slightly better. The difference in performance was significant for the reconstruction of whole plant and became less significant as we moved to the reconstruction of the leaf nodes. These results show that the reconstruction performance of the NBV planner on complex plants is almost as efficient as with simpler plants, especially when reconstructing finer plant parts such as leaf nodes. The performance of the attention-driven NBV planner was not very sensitive to the amount of occlusion, which implies that it can be used effectively even in highly occluded scenarios.

**Number of candidate viewpoints**. Varying the number of candidate viewpoints in the NBV algorithm did not have a significant influence on the results, as shown in Fig. 15. This observation deviated from our expectation that a higher number of candidate viewpoints could lead to a higher accuracy in less views. This deviation might be due to a couple of reasons. First, the restriction imposed by the cylindrical-sector constraint forced all the candidate viewpoints to be centred along the main stem. Hence, even when a higher number of candidate viewpoints were sampled, the captured information was not very different from the previous candidates. Second, even with a few candidate viewpoints, the pseudo-random sampling strategy (Section 2.2, Stage II) ensured that all the viewpoints were evenly spaced across the plant. When viewpoints are sampled carefully and evenly across the plant under the cylindrical-sector constraint, a low number of candidate viewpoints seems to be sufficient to reap the benefits of an attention mechanism. A lower number of candidate viewpoints is also favourable since the computation time is reduced. When the cylindrical-sector constraint is relaxed or the pseudo-random sampling strategy is removed, we still expect that a higher number of candidate viewpoints will lead to higher accuracy in less views.

**Resolution of reconstruction**. When the resolution of the OctoMap was reduced, the performance of the NBV planner slightly improved, as shown in Fig. 16. The improvement in performance can be attributed to the increase in the distance threshold $\rho$ of the F1-score metric. When using voxel resolutions of 0.005 m and 0.007 m, the distance threshold was equated to the voxel resolution and hence increased to 0.005 m and 0.007 m respectively. In effect, for lower resolutions, the evaluation criterion was less critical. The improvement in performance was significant for the reconstruction of the whole plant and became less significant as we moved to the reconstruction of the leaf nodes. The results imply that the performance of the NBV planner on higher resolutions is almost as efficient as with lower resolutions. The performance of the attention-driven NBV planner was not very sensitive to minor changes in voxel resolution and can be used effectively to reconstruct both coarser and finer details of the plant depending on the task-at-hand.

### 4.2. Real-world results

We report the performance of the NBV planners on real-world tomato plants in a greenhouse on the task of reconstructing the plant nodes. We observed that the NBV-leaf-nodes planner significantly outperformed the other planners, as shown by the F1-scores in Fig. 17a. We performed a Kruskal-Wallis H Test to determine the statistical significance of the NBV-leaf-nodes planner. It is illustrated by the asterisks in Fig. 17, which indicate the viewpoints for which the NBV-leaf-nodes planner was statistically significant than all the other with a p-value less than 0.05. Fig. 17b and c show the precision and recall performances (calculated according to Equations (7) and (8)), which clearly indicate that all planners suffered from poor precision. The NBV-leaf-nodes planner had a precision of 52.8% at the end of 3 views, while the NBV-whole-plant, predefined, and random planners had an equally low precision of 53.9%, 49.9%, and 53.9% respectively. On the other hand, in terms of recall, the NBV-leaf-nodes planners significantly outperformed the other planners (Fig. 17c). After 3 views, the NBV-leaf-nodes planner had a recall of 57.8%, while the NBV-whole-plant, predefined, and random planners had a recall of 43.1%, 31.3%, and 42.4% respectively. The recall results indicate that the NBV-leaf-nodes planner was able to efficiently visit informative viewpoints for reconstruction, but the performance in terms of F1-score was limited due to low precision.

The reason for low precision was likely due to missing points in the ground-truth point cloud and noise in the reconstructed point cloud. The ground-truth point cloud obtained from COLMAP required colour images of the scene from multiple views to estimate the depth. So, parts of the scene that were only visible from one view were not reconstructed. The reconstructed point cloud, however, directly used measurements from the depth sensor and could still reconstruct parts of the scene that were only visible from one view. This discrepancy led to a lower precision, even though the reconstructed points could have been correct. The reconstructed point cloud also had some noise due to sensor noise and errors in camera poses, which again contributed to the low precision. From Fig. 17b, we observe that there were no significant differences in the precision performances and all planners suffered equally due to missing ground-truth points and noisy reconstruction points.

For comparing the speed of reconstruction, we reduced the accuracy threshold $\tau_a$ on the F1-score to 60%, since all the planners had an accuracy lower than 70%. The NBV-leaf-nodes planner reached $\tau_a = 60\%$ in 5 viewpoints, 2 views faster than the NBV-whole-plant planner. The random and predefined planners did not reach $\tau_a = 60\%$ within ten viewpoints. After 3 views, the NBV-leaf-nodes planner reconstructed the nodes 6.5%, 17.3%, and 7.7% more than the NBV-whole-plant, predefined, and random planners respectively in terms of their F1-score.

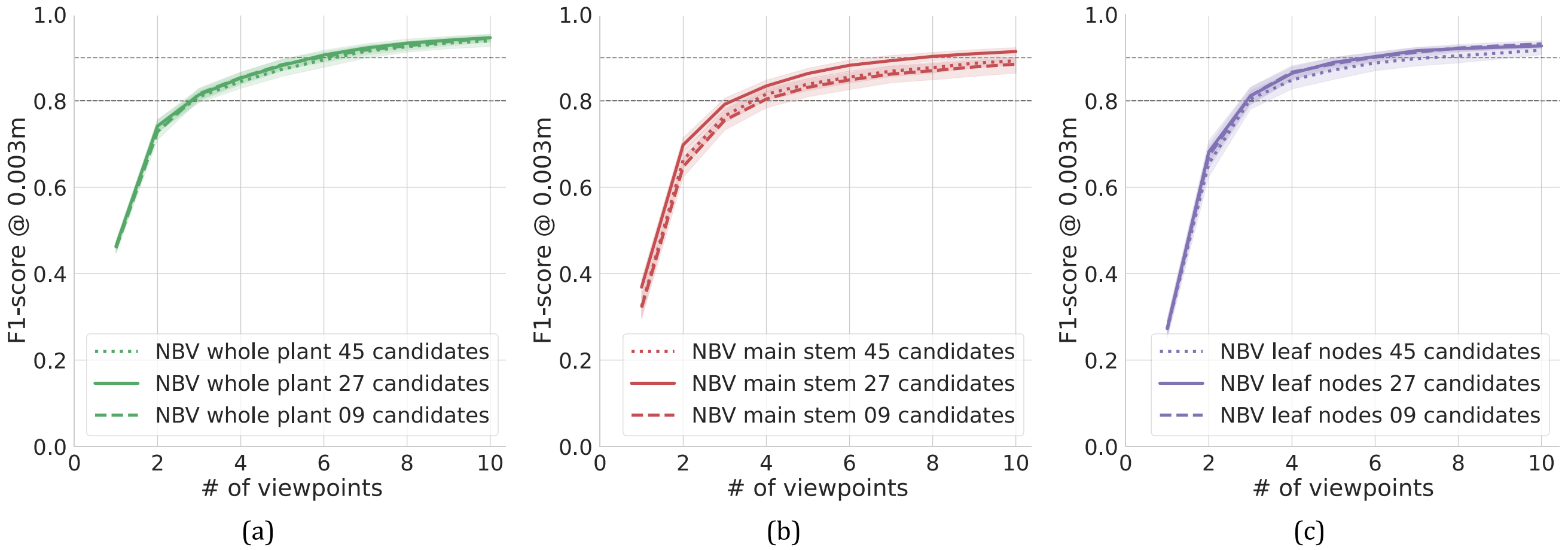

**Fig. 15.** Sensitivity of the performance of attention-driven NBV planner to changes in the number of candidate viewpoints from 27 views to 9 and 45 views. The plots show the comparison of F1-scores at different number of candidate viewpoints for the reconstruction of (a) the whole plant, (b) the main stem, and (c) the leaf nodes. The shaded regions represent the 95% confidence interval of the mean.

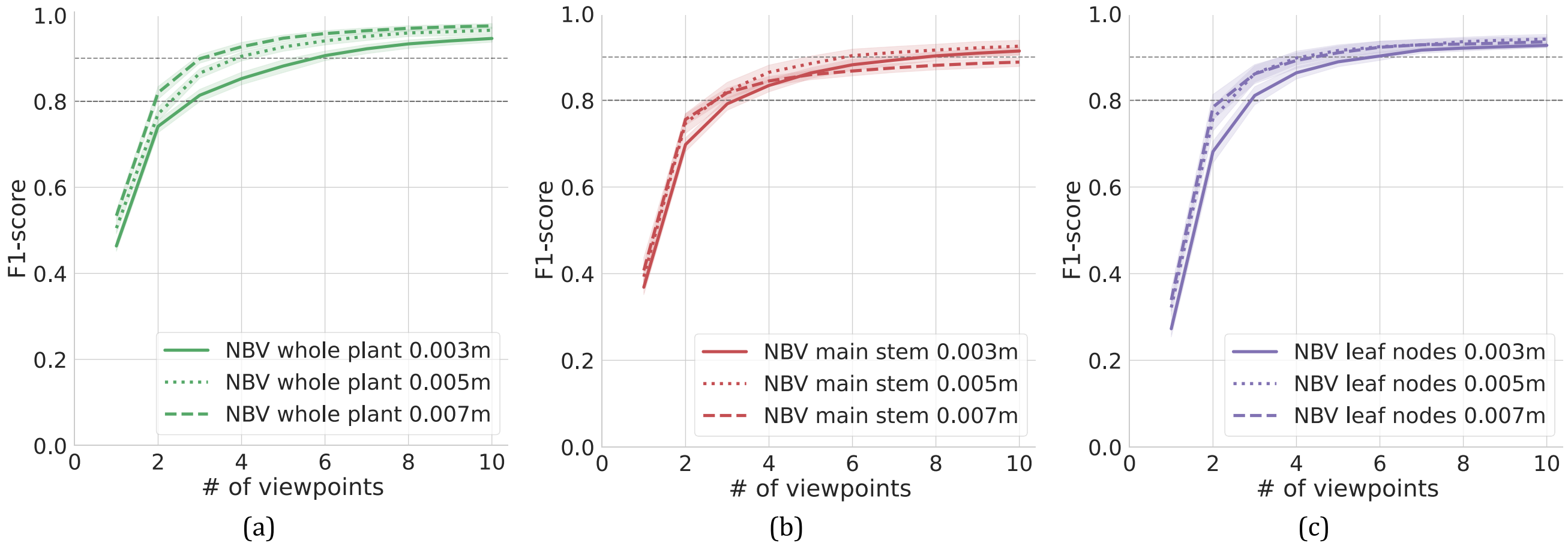

**Fig. 16.** Sensitivity of the performance of attention-driven NBV planner to changes in voxel resolution from 0.003 m to 0.005 m and 0.007 m. The plots show the comparison of F1-scores at different voxel resolutions for the reconstruction of (a) the whole plant, (b) the main stem, and (c) the leaf nodes. The shaded regions represent the 95% confidence interval of the mean.

Fig. 17d shows the distribution density of the F1-score after five views over all 84 experiments. From this plot, we observe that the NBV-leaf-node planner's performance was more repeatable as the F1-score density had low spread and had a good accuracy with a median F1-score of 61.6%, i.e., half the experiments had an F1-score greater than this. The performance of the NBV-whole-plant planner was slightly less repeatable and accurate with a median F1-score of 59.3%. The predefined and random planners were least repeatable as indicated by a large spread and were less accurate with median F1-scores of 52.8% and 58.5% respectively after five views. These results indicate that the attention mechanism consistently outperformed the other planners across all experiments.

The real-world results support the simulation results and further indicate the significance of an attention mechanism for fast and accurate reconstruction of the plant nodes. Our results are valid under complex real-world greenhouse conditions, such as natural variation and occlusion in tomato plants, natural illumination, sensor noise, and uncertainty in camera poses.

#### *4.2.1. Computation time and memory utilisation*

The average time required to compute the next-best viewpoint was 0.47 s, using a 2.60 GHz Intel Core i7-9750H CPU. Since we sampled 30 candidate viewpoints, this meant that the average time to compute the expected information gain of one candidate viewpoint was roughly 0.016 s. We applied some optimisation techniques to achieve these high computation speeds. First, we only evaluated the rays that intersected with the ROIs. Else, the rays were ignored, since they would not contribute to the expected information gain. This significantly reduced the total number of rays that needed to be processed. Second, we parallelly estimated the information gains for multiple rays using OpenMP (Dagum & Menon, 1998), which further improved the computation speed. Without these modifications, the computational complexity of choosing the next-best viewpoint would be $\mathcal{O}(n)$, where n is the number of candidate viewpoints. Hence, the computation time will increase linearly with the number of candidates. However, in our case, this complexity was significantly reduced due to the above modifications. The total memory usage of our algorithm was roughly 839 MB with a CPU utilisation of 8.5% across 12 cores.

## 5. Discussion

Our results show that an attention mechanism can be used to efficiently guide active vision towards the relevant plant parts according to the task-at-hand. For 3D reconstruction of the whole plant, main stem, and leaf nodes, the NBV planner performed best when an appropriate region of interest was provided. Our results, in general, agree with the work of Isler et al. (2016), who found that using object-focused formulations of volumetric information improved the performance of 3D reconstruction compared to object-agnostic formulations. Our results are also in line with the work of Zaenker et al. (2021), who observed a gain in perception of fruits when labelled regions of interest were provided to an NBV planner. As an addition, in this paper, we clearly showed the advantage of adding an attention mechanism in terms of significant gains in accuracy and speed of 3D reconstruction, and we validated the results using data of real-world tomato plants in a greenhouse.

### *5.1. Errors in simulation and real-world experiments*

In the simulation experiments, the conditions were ideal with no sensor noise, no errors in camera poses, and no changes to illumination or background. The challenge for the algorithm was to deal with variation and occlusion caused by the tomato plants of varying growth stages and structural complexity. Hence, the simulation results clearly indicate how well the attention-driven NBV planner can deal with variation and occlusion, when the sources of error are controlled.

In the real-world experiments, apart from the natural variation and occlusion of plants, the robot was exposed to different real-world complexities. First, the plants were mainly illuminated by ambient light from outside and partially by LED strip lights in the background. The illumination varied as the lighting conditions outside the greenhouse varied throughout the day. Second, the depth measurements by the Realsense L515 camera was noisy. As the camera used a LiDAR-based depth sensor, it was easily influenced by the ambient light and the LED strip lights, which led to noisy depth measurements. Finally, there were errors in the camera poses, potentially originating from the unevenness of the platform on which the robot was mounted and calibration error between the robotic arm and the camera. None of these conditions were controlled for our experiments.

The real-world complexities can lead to a noisy 3D reconstruction of the plant, consisting of false positive points (i.e. reconstructing points in space that should be empty) and false negative points (i.e. failing to reconstruct some portions of the plant). However, since we combined the sensor information from multiple viewpoints in a probabilistic way using an OctoMap, these errors were largely mitigated. Moreover, the attention mechanism proposed in this paper helped the NBV planner to focus only on the relevant plant parts, further reducing the influence of noise on the viewpoint planning. Hence, we believe that the use of OctoMap and the attention mechanism led to efficient and reliable performance of the NBV planner even in real-world conditions. Further improvements can be made by using more precise eye-in-hand calibration techniques, stabilising the robot platform, estimating the camera poses using visual odometry methods, and applying advanced sensor fusion techniques to further reduce the noise in depth measurements.

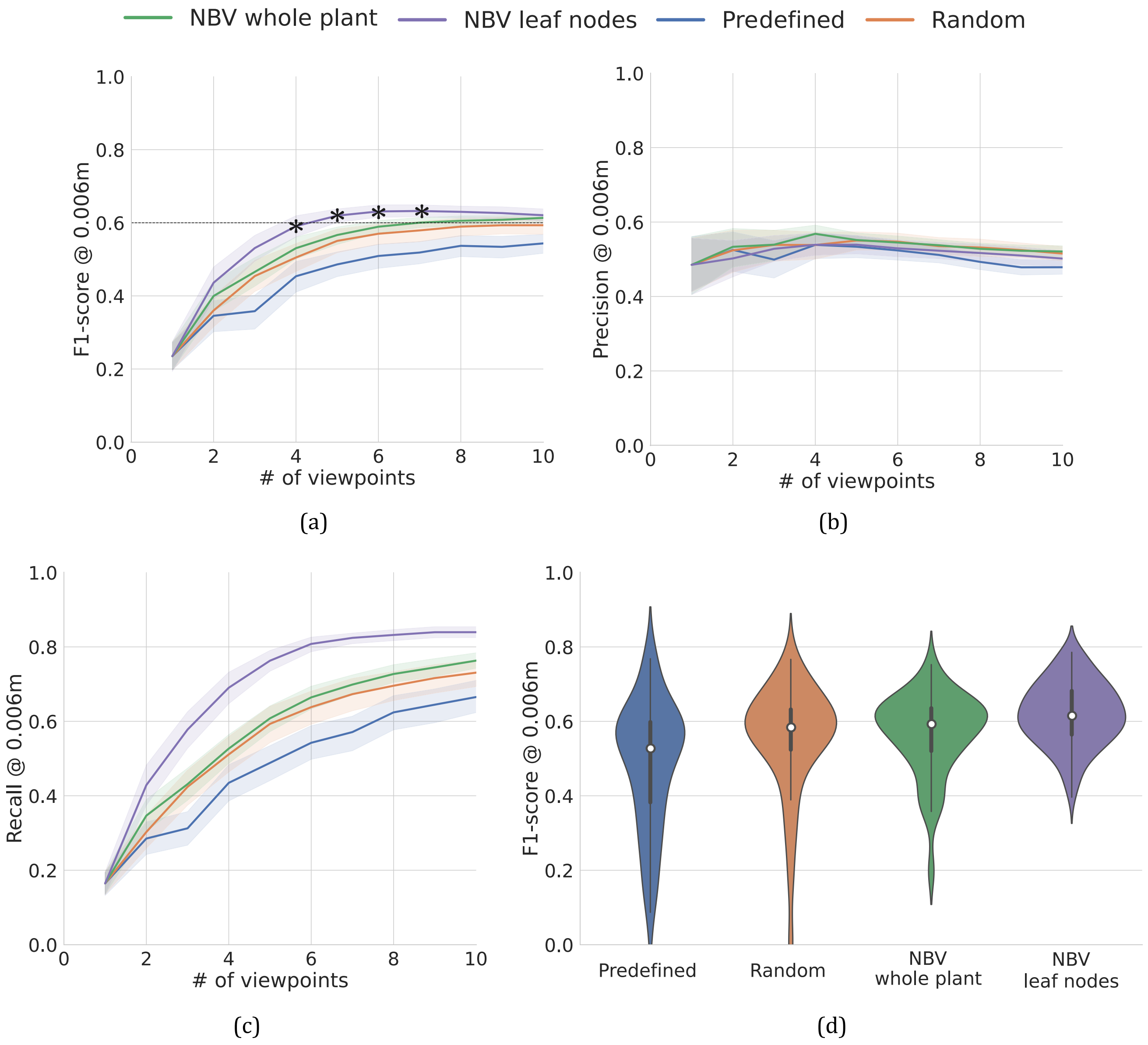

**Fig. 17.** Real-world performance of the planning algorithms on the reconstruction of the plant nodes for tomato plants in a greenhouse. (a) The F1-score, (b) precision, and (c) recall of the planners over ten viewpoints. The shaded regions represent the 95% confidence interval of the mean over 84 experiments. The asterisks (*) denote the viewpoints for which the best-performing planner was significantly better than all the other planners with a p-value less than 0.05. (d) The distribution density plot of the F1-score for the planners after five views over 84 real-world experiments. The median F1-scores are shown by the white dots.

### 5.2. Constraint on viewpoint sampling

We added a cylindrical-sector constraint on viewpoint sampling, which greatly benefited the random planner, since any sampled viewpoint was always oriented towards the centre of the plant. This made its performance comparable to that of the predefined and NBV planners. Moreover, it also led to insignificant differences in the performance of NBV-whole-plant and NBV-main-stem in simulation. However, such a constraint was necessary to sample reasonable viewpoints that would be oriented towards the plant and be reachable by the robotic arm. Under the cylindrical-sector constraint, the attention-driven NBV planner provided significant gains in speed and accuracy. We believe that in a more unconstrained setting, the gains of an attention-driven NBV planner will be even greater. In the absence of any constraint, the sampled viewpoints can be positioned and oriented arbitrarily, which means most of the viewpoints might not even have the plant in view. To ensure that there are a reasonable number of viewpoints that have the plant in view, a large number of viewpoints need to be sampled, which might be computationally infeasible. An attention-driven NBV planner will still be able to pick viewpoints that are informative and have the relevant plant parts in view, whereas a random planner is expected to perform poorly. Future work could focus on studying the impact of relaxing constraints for viewpoint sampling.

In the real-world experiments, we constrained the view selection to 100 discrete prerecorded viewpoints to ensure comparability and repeatability of our experiments. This limited the performance of the NBV planner. If the NBV planner is allowed to plan viewpoints in continuous space, it can potentially choose better viewpoints with higher information gains, which would improve the reconstruction performance.

### 5.3. Defining regions of interest

The ROIs for the attention mechanism were provided manually in our experiments. However, for real-time operation of NBV planning in practice, the algorithm should be able to automatically define the ROIs. The performance of the planners with manually-defined ROIs, shown in this paper, can be considered as an upper limit, which a similar algorithm with automatically-defined ROIs can aim to achieve. One approach to automatically defining the ROIs is to train convolutional neural networks to detect plant parts (Barth et al., 2019; Boogaard et al., 2020, 2023; Koirala et al., 2019) and define the ROIs around the detections according to the task-at-hand. A more promising approach is to reason about and predict ROIs based on current detections and prior knowledge about the plant structure. Since most commercial crops such as tomatoes are grown in a structured way, prior knowledge about their growth structure can be used to reason about ROIs. Such knowledge can be used to predict where the robot can expect to see another plant node and guide active vision. Future work could investigate these options.

### 5.4. Evaluation of real-world experiments

A perfect ground truth for real-world experiments does not exist. The reconstruction from COLMAP, used as ground truth in this study, was not perfect and had some missing information due to occlusions (as discussed in Section 4.2), which was not ideal for evaluation. Further, COLMAP reconstruction is also affected by the quality of colour images, errors in feature detection and feature matching, and errors in the camera poses that were provided. However, we believe that the imperfection in our ground truth had a similar effect on all the planners and hence their performances were comparable. Given a perfect ground truth, we believe that the performances of the planners will follow a similar trend as our results, with the attention-driven NBV planner significantly outperforming the other planners.

For the real-world experiments, we used a row of tomato plants that were relatively singulated from one other, although occlusions were caused by leaves of neighbouring plants. This is the level of occlusion that is commonly expected in a commercial greenhouse. However, in some rare cases, it is possible that some neighbouring plants are too close to each other, and their leaves and stems are intertwined. We expect that this added complexity will only support the need for an attention mechanism even more. As shown in Section 4.1.4, the attention-driven NBV planner performs effectively even under higher levels of occlusions. Future work could focus on evaluating the advantage of an attention mechanism with such complex situations.

### 5.5. Criteria for evaluation

Speed, in terms of execution time, is an important criterion for most real-world applications of active vision. However, we did not use execution time as an evaluation metric in this work because it depends largely on the robotic agent used and, in some cases, it can be reduced simply by optimising or upgrading the setup (e.g. GPU acceleration). We chose to quantify speed based on the number of viewpoints since it is agent-agnostic and provides a more overarching insight about the performance of the planners. Nevertheless, for deployment in the real world, it might be necessary to take the overall execution time of the NBV planner into account. When the execution time is added as a constraint for view selection, there will be a trade-off between the information gain and execution time. The performances of the planners might look different than what we presented in this paper, but we expect that the attention-driven NBV planner will still outperform the other planners. Future work could test this hypothesis by taking execution time into consideration for view selection and evaluating its impact.

## 6. Conclusion

The objective of this paper was to investigate the gain in accuracy and speed of 3D reconstruction of plants and targeted plant parts when an attention mechanism is added to next-best-view (NBV) planning methods. Using simulation and real-world experiments, it was shown that an attention-driven NBV planner significantly outperforms predefined and random planning strategies at all levels of attention when appropriate regions of interest are provided. In simulation, the attention-driven NBV planner reconstructed 80% of the whole plant with 1 less view compared to predefined and random planners and reconstructed the whole plant 9.7% and 5.3% more respectively after 3 views. Similarly, the attention-driven NBV planner reconstructed 80% of the main stem with 1 less view compared to predefined and random planners and reconstructed the main stem 14.2% and 7.9% more respectively after 3 views. For finer perception targets, such as leaf nodes, the advantage of an attention mechanism was more pronounced. The attention-driven NBV planner reconstructed 80% of the leaf nodes with 3 less views compared to predefined and random planners and reconstructed the leaf nodes 25.9% and 17.3% more respectively after 3 views. The results of this work indicate that providing the appropriate region of interest according to the task-at-hand is important for reconstruction performance. It was shown that the attention-driven NBV planner works effectively despite changes to the plant models, the amount of occlusion, the number of candidate viewpoints, and the resolutions of reconstruction.

To the best of our knowledge, this work was the first to test the advantage of an attention mechanism for NBV planning using a robotic setup in a real-world greenhouse environment. The simulation results were confirmed by real-world experiments, where natural variation and occlusion in tomato plants, natural illumination, sensor noise, and uncertainty in camera poses majorly influenced the viewpoint planners. Here, the attention-driven NBV planner reconstructed 60% of the nodes within 5 views, whereas the predefined and random planners did not reach an accuracy of 60%. It reconstructed the nodes 17.3% and 7.7% more than the predefined and random planners respectively after 3 views.

It could be concluded that an attention mechanism for NBV planning is necessary for targeted perception in complex agro-food environments. We hope this work convinces researchers and practitioners in agro-food to see the benefits of NBV planning and adopt it over fixed or predefined strategies for perception. Given that an attention mechanism is advantageous, future work can focus on better strategies for view planning that can intelligently assign ROIs based on scene understanding, consider the execution time for view selection, or plan multiple steps ahead. Future experiments in real-world environments can study the impact of reducing the constraint on view sampling and test on special cases such as extreme levels of occlusion.

## Funding

This research is funded by the Netherlands Organization for Scientific Research (NWO) project Cognitive Robots for Flexible Agro-Food Technology (FlexCRAFT), grant P17-01.

## CRediT authorship contribution statement

**Akshay K. Burusa:** Writing – original draft, Software, Methodology, Investigation, Data curation, Conceptualization. **Eldert J. van Henten:**

Writing – review & editing, Supervision, Funding acquisition, Conceptualization. **Gert Kootstra:** Writing – review & editing, Supervision, Funding acquisition, Conceptualization.

## Declaration of competing interest

The authors declare that they have no known competing financial interests or personal relationships that could have appeared to influence the work reported in this paper.

## Acknowledgements

We thank Dide van Teeffelen and Bolai Xin for contributing towards a preliminary study. We thank David Rapado-Rincón for assisting in collecting real-world data. We thank the members of the FlexCRAFT project for engaging in fruitful discussions and providing valuable feedback to this work.